\documentclass[journal,twoside,web]{ieeecolor}
\usepackage{generic}
\let\labelindent\relax
\usepackage{enumitem}
\usepackage{cite}
\usepackage{amsmath,amssymb,amsfonts}
\usepackage{algorithmic}
\usepackage{graphicx}
\usepackage{textcomp}
\usepackage{amsfonts}
\usepackage{subcaption,subfloat}
\usepackage{multirow}
\usepackage{tikz,pgfplots,pgfplotstable}   
\usepackage{booktabs}
\usepackage{hyperref}
\usepackage{makecell}
\usepackage{nicefrac}
\definecolor{forestgreen}{rgb}{0.13, 0.55, 0.13}
\usepgfplotslibrary{groupplots,fillbetween,colorbrewer,statistics}
\usetikzlibrary{intersections,calc,arrows,matrix,spy,pgfplots.statistics, pgfplots.colorbrewer}
\pgfplotsset{plot coordinates/math parser=false}
\usepackage{color}
\definecolor{darkred}{rgb}{0.7,0,0}
\definecolor{darkgreen}{rgb}{0,0.5,0}
\definecolor{darkblue}{rgb}{0,0,0.7}
\definecolor{SkyBlue}{rgb}{0.53, 0.81, 0.92}

\def\BibTeX{{\rm B\kern-.05em{\sc i\kern-.025em b}\kern-.08em
    T\kern-.1667em\lower.7ex\hbox{E}\kern-.125emX}}
\usepackage{pifont}
\newcommand{\cmark}{\ding{51}}%
\newcommand{\xmark}{\ding{55}}%
\setlist[itemize]{leftmargin=*}

\begin{document}
\bstctlcite{IEEEexample:BSTcontrol}
\title{Towards a Comprehensive Solution for a Vision-based Digitized Neurological Examination}
\author{Trung-Hieu Hoang$^*$, Mona Zehni$^*$\thanks{$^*$ Both authors contributed equally.}, Huaijin Xu, George Heintz, Christopher Zallek, Minh N. Do
\thanks{
Trung-Hieu Hoang, Mona Zehni and Minh N. Do are with the Department of Electrical \& Computer Engineering, and Coordinated Science Laboratory at University of Illinois at Urbana-Champaign (UIUC). Minh N. Do is also with the VinUni-Illinois Smart Health Center, VinUniversity, Hanoi, Vietnam. Huaijin Xu is with the Department of Kinesiology \& Community Health at UIUC.
George Heintz is with Healthcare Engineering Systems Center at UIUC.
Christopher Zallek is with OSF HealthCare Illinois Neurological Institute - Neurology.
}
\thanks{This article has been accepted for publication in a future issue of the IEEE Journal of Biomedical and Health Informatics, but has not been fully edited. Content may change prior to final publication. Citation information: DOI 10.1109/JBHI.2022.3167927, IEEE Journal of Biomedical and Health Informatics.}
}

\maketitle

\begin{abstract}
The ability to use digitally recorded and quantified neurological exam information is important to help healthcare systems deliver better care, in-person and via telehealth, as they compensate for a growing shortage of neurologists. Current neurological digital biomarker pipelines, however, are narrowed down to a specific neurological exam component or applied for assessing specific conditions. In this paper, we propose an accessible vision-based exam and documentation solution called Digitized Neurological Examination (DNE) to expand exam biomarker recording options and clinical applications using a smartphone/tablet. 
Through our DNE software, healthcare providers in clinical settings and people at home are enabled to video capture an examination while performing instructed neurological tests, including finger tapping, finger to finger, forearm roll, and stand-up and walk. Our modular design of the DNE software supports integrations of additional tests. The DNE extracts from the recorded examinations the 2D/3D human-body pose and quantifies kinematic and spatio-temporal features. The features are clinically relevant and allow clinicians to document and observe the quantified movements and the changes of these metrics over time. A web server and a user interface for recordings viewing and feature visualizations are available. 
DNE was evaluated on a collected dataset of 21 subjects containing normal and simulated-impaired movements. The overall accuracy of DNE is demonstrated by classifying the recorded movements using various machine learning models. Our tests show an accuracy beyond 90\% for upper-limb tests and 80\% for the stand-up and walk tests.

\end{abstract}
\vspace{-3pt}

\vspace{-10pt}
\section{Introduction}
\label{sec:intro}
\vspace{-5pt}
The burden and prevalence of neurological disorders~\cite{Feigin_2021} and the national shortage of neurologists~\cite{Dall_2013} continue to grow hand in hand. This increases disparity through unequal access to clinical care and drives worsening clinician burnout rates.
Meanwhile, the COVID-19 pandemic has boosted the transition from in-person to virtual neurological examinations~\cite{grossman_rapid_2020, al_hussona_virtual_2020} through teleneurology (TN) platforms. Raplidly developing TN has shown potential in making efficient assessments remotely~\cite{capozzo_telemedicine_2020, duncan_video_2020,patterson_neurological_2021} and helping in distributing scarce healthcare resources and enhancing accessibility to neurological care~\cite{Mutgi_2015, Dorsey_2018}.
In addition, digital biomarker exam
solutions with quantification of physical evaluations that bypass clinician availability and subjectivity of assessments~\cite{Adams2021} are important to improve care and compensate for the shortage of neurologists. 

Current digital biomarker exam systems are devoted to a single neurological test~\cite{Williams2020,pmlr-v68-jaroensri17a,Xue2018}, require advanced setups/equipment~\cite{tedim_cruz_novel_2014}, or lack automated assessments~\cite{einsler_sarahome_2021},~\cite{Wei2021}. Therefore, a digital biomarker solution, 1) suitable for use by neurologists and non-neurologists, 2) with wide applicability at clinics or home, 3) that is easy to deploy, 4) supports a wide range of neurological tests, and 5) enables automated objective quantitative evaluations, would significantly advance health care delivery.

\begin{figure}[t!]
    \centering
    \includegraphics[width=0.95 \linewidth]{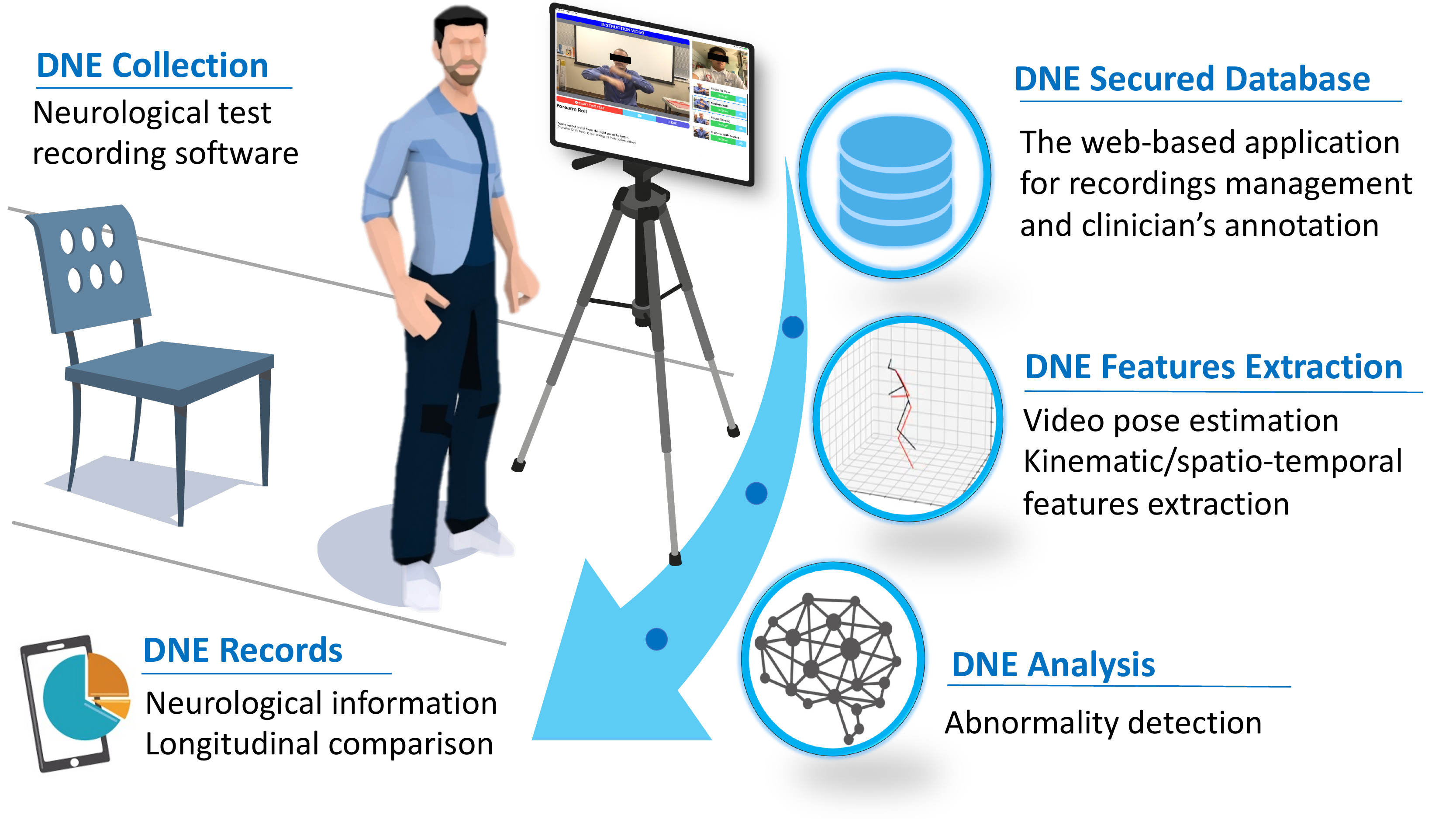}
    \vspace{-5pt}
    \caption{Illustration of our digitized neurological exam system. 
    }
    \label{fig:dne_framework}
\end{figure}

For this purpose, in this work, we introduce an end-to-end vision-based exam and documentation platform named Digitized Neurological Examination (DNE).
As part of DNE, we designed an easy-to-use smartphone/tablet software with pre-defined examination instructions. The DNE software allows the users to video record their performance on several neurological screening examinations, including finger tapping (FT), finger to finger (FTF), forearm roll (FR), and stand-up and walk (SAW). 
These recordings are uploaded to a secure cloud-based storage. In an offline step, for each recording, 2D/3D pose, estimating the location of major human body keypoints is extracted using deep-learning-based solutions such as OpenPose~\cite{Cao2019_openpose}, and VideoPose3D~\cite{pavllo2019_videopose3d}. From the estimated pose, unified digital biomarkers, including spatio-temporal and kinematic features, are computed \cite{john_biomechanics}. We showcase the performance of our system on a dataset collected from $21$ healthy subjects taking different neurological tests (FT, FTF, FR, SAW) when their function is normal or with a simulated impairment. We incorporate our defined features in a variety of machine learning models to detect abnormal functioning in our dataset. Fig.~\ref{fig:dne_framework} illustrates the capabilities our DNE system. 

We summarize the key contributions of this work as: 
\begin{itemize}
    \item We develop a unified and modular software package for high-quality DNE recording collection. Our DNE software is easy-to-use, allows the integration of new tests, and runs on handheld iOS devices. We also implement a web-based dashboard for viewing the recordings and feature visualization. 
    \item We propose a vision-based approach to study various neurological tests (FT, FTF, FR, and SAW). For each test, we define clinically interpretable kinematic and spatio-temporal quantified features. 
    \item To the best of our knowledge, we are the first to construct a vision-based dataset consisting of multiple neurological tests and simulated-impaired video recordings per subject alongside the extracted 2D/3D pose. Analyzing this dataset allows us to have a normal self-baseline for each abnormal recording and test the power of the extracted features in distinguishing normal from abnormal performance. Our dataset (excluding RGB videos due to privacy restrictions) and code will be available at \url{https://dneproject.web.illinois.edu/}.
\end{itemize}

The organization of this paper is as follows. Section \ref{sec:related_work} summarizes recent studies on digital biomarker  systems. Section \ref{sec:system_design} describes DNE's software platform used in our data collection. Section \ref{sec:dataset_collection} introduces our DNE dataset. We define our features in detail in Section \ref{sec:vision_analysis}. Section \ref{sec:results} contains our analysis results while Section \ref{sec:conclusion} draws our main conclusions.

\vspace{-10pt}
\section{Related Work}
\vspace{-5pt}
\label{sec:related_work}
\begin{figure*}[t!]
    \hspace*{-1.cm}
    \begin{subfigure}[b]{\columnwidth}
    \centering
    \includegraphics[width=0.65\columnwidth]{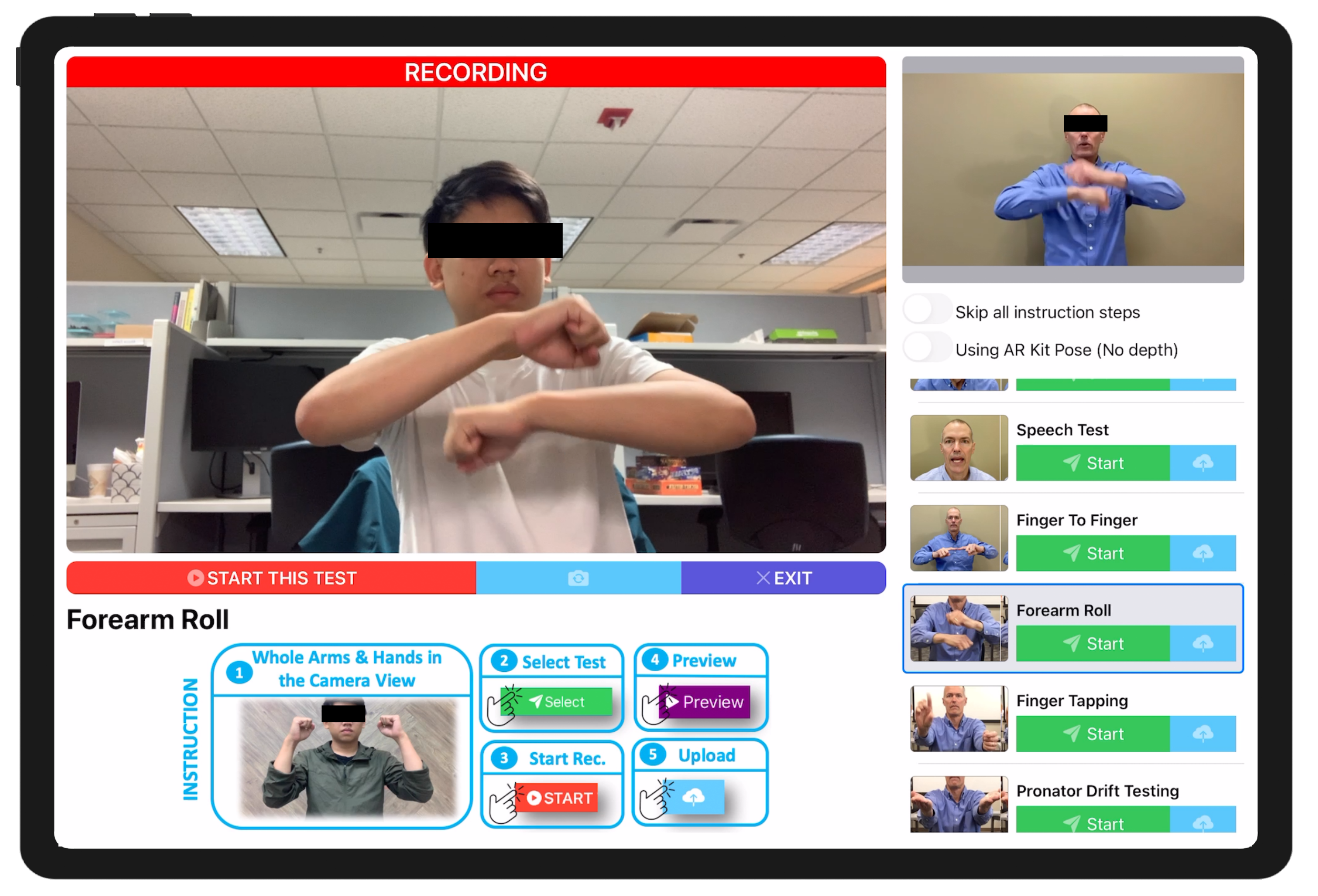} 
    \caption{DNE Recorder}
    \label{fig:dne-recorder}
    \end{subfigure}
    \hspace*{-1.3cm}
    \begin{subfigure}[b]{\columnwidth}
    \centering
    \includegraphics[width=1.3\columnwidth]{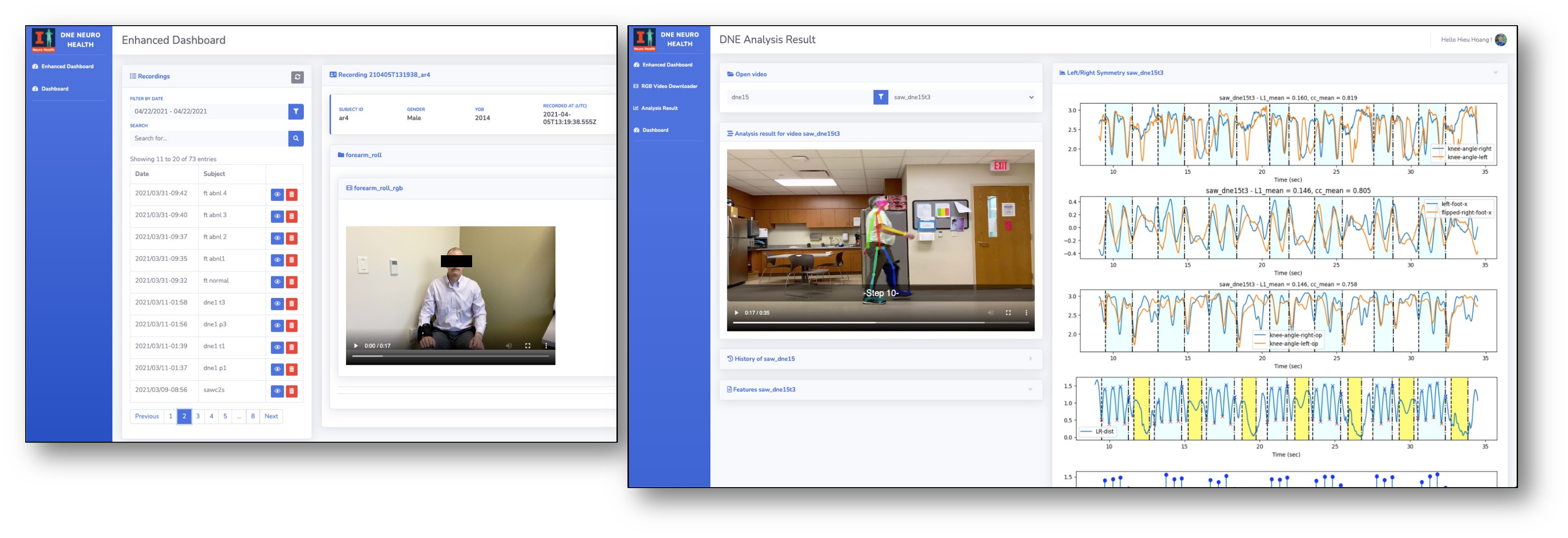}
    \caption{DNE Viewer}
    \label{fig:dne-viewer}
    \end{subfigure}
    \caption{DNE System. (a) \textit{DNE Recorder} - an iOS application for neurological recordings collection. (b) \textit{DNE Viewer} a web application for dataset management, video previewing and visualizing the analysis results (best viewed in magnification).}
\end{figure*}

In this section, we review the related literature to different tests  (FT, FTF, FR, SAW). For each test, we briefly discuss the existing sensor, web/smartphone and vision-based solutions.

\noindent\textbf{Finger Tapping (FT):} 
Sensor-based FT assessments study spectral analysis of gyroscope data~\cite{jovicic2014}, opening finger tap velocity captured by accelerometers~\cite{Yokoe2009}, standard deviation, range and entropy measured by a collection of sensors including synchronized wrist watches, pressure sensors and accelerometers~\cite{Jia2014}. Several smartphone based applications~\cite{Lee2016,Memedi2015,Aghanavesi2017,Zhang2018} are designed to quantitatively evaluate various symptoms and motor skills in patients with Parkinsons Disease (PD). While these approaches are proven effective and low cost, their measurements are not as informative as vision-based methods, relying on video data and simulating in-person clinical examinations. 
Among vision-based pipelines, \cite{Jobbagy2005,Khan2014,Liu2019,Williams2020} extract a set of kinematic interpretable features from the tracked positions of the fingers given an RGB video. These features are easy to explain and associate with clinical symptoms. On the other hand, black box deep learning models operating on the estimated finger poses and their derivatives are proposed in~\cite{Li2021}. While these solutions provide high accuracy, unlike our DNE, they lack explainability and require large training sets to generalize and avoid overfitting.

\noindent\textbf{Finger to Finger (FTF):} A well-studied test in the literature that is similar to FTF in terms of measuring smoothness and upper extremity coordination is the finger to nose test. 
Among sensor-based methods, Rodrigues \textit{et al.} in \cite{rodrigues_chronic_stoke_2017} investigates the coordination ability of patients with chronic stroke versus healthy control using a complex marker-based motion analysis system. Oubre \textit{et al.}~\cite{oubre_decomposition_2021} studied ataxia through wearable inertial sensors and a computer tablet version of finger to nose test. Furthermore, predicting severity levels of ataxia or PD via a rapid web-based computer mouse test is explored in~\cite{gajos_computer_mouse_2020}. 
Jaroensri \textit{et al.} \cite{pmlr-v68-jaroensri17a} is among the first to propose vision-based solutions that are on par with a specialist in terms of rating the severity scale of PD while relying on estimated joint positions from recorded videos.

\noindent\textbf{Upper Limb Tests:} 
To the best of our knowledge, sensor-based or vision-based studies related to the forearm roll task are scarce. Thus, here we further overview the existing methods devoted to the study of upper limb movements.
Using wearable sensors, Cruz \textit{et al.} in~\cite{tedim_cruz_novel_2014} assessed the acceleration, velocity or smoothness of the upper limb motor function of patients after stroke. A low-cost Kinect based solution, tracking subjects' hand when asked to move a marker on a rectangular pattern is proposed in~\cite{Simonsen2017}. The range of motion is analyzed using an internet-based goniometer in~\cite{Hoffmann_2007}. 
In~\cite{allin_robust_2010}, the authors describe a vision-based system that captures upper limb motions via multiple cameras installed at different views. While this multi-camera system is less sensitive to occlusions and dynamic backgrounds, unlike our DNE system, it requires a special setup which is hard to install for home-use.

\noindent\textbf{Stand-up and Walk (SAW):} In our review of gait analysis literature, we focus on the marker-less~\cite{kour_computer-vision_2019} vision-based solutions, mainly measured using general handheld cameras and mobile devices. 
In early efforts for marker-less gait analysis, silhouettes are extensively used to detect heel-strike and toe-off occurrences. These two events refer to the first and last ground contact of each foot, later on adopted to accurately estimate important gait parameters \cite{ortells_vision-based_2018, nieto-hidalgo_gait_2018, zhu_computer_2016,NIETOHIDALGO2016}. However, these methods are restricted to specific laboratory settings and are sensitive to the quality of foreground/background segmentation. The surge of research in the human pose estimation field~\cite{Toshev2013,Sun_2019_CVPR,Martinez_2017_ICCV} brought along popular deep learning frameworks which accurately estimate the 2D/3D location of body joints from different inputs including RGB image, video and depth maps~\cite{Cao2019_openpose, Rong2020, Zimmermann2018, pavllo2019_videopose3d}. Depth-map based gait assessment solutions relying on the estimated pose from either depth or RGBD~\cite{Clark2018,Springer2016}, have studied the rotational angle and angle velocity of certain body keypoints~\cite{Dolatabadi2016} and evaluated the spatio-temporal gait metrics such as step length and time~\cite{Xue2018, Joao2020}.

Wei \textit{et al.~} \cite{Wei2021} introduced an automated smart-phone based video capturing system with hand/body pose estimation. While neurological exams such as gait are considered in~\cite{Wei2021}, feature extraction and analysis is not studied and the main focus is on the quality control of the video acquisition
process. Using the estimated pose from OpenPose~\cite{Cao2019_openpose}, Xue \textit{et al.} \cite{Xue2018} studied the remote monitoring of gait parameters for senior care. Furthermore,~\cite{Li2018NS} reports timings of different segments of the timed-up-and-go (TUG) test by performing frame-based activity classification based on 2D pose data. To assess the freezing of gait (FoG) symptom in Parkinson patients,~\cite{sato_quantifying_2019} proposed the use of frequency analysis methods while~\cite{Hu2020} adopted graph convolutional neural networks to attain the probability of FoG from pose data. Kidzi{\'n}ski in~\cite{kidzinski_deep_2020} employed black-box deep learning models to estimate the level of movement disorder in children suffering from cerebral palsy. Despite their promising results, deep learning based solutions are less interpretable and require large training supervised datasets for better generalization.

\vspace{-10pt}
\section{System Design}
\vspace{-5pt}
\label{sec:system_design}

As part of DNE, we developed three software packages to maintain data acquisition, analysis and results report.

\noindent \textbf{DNE Recorder:} 
This module accommodates easy-to-use self or assisted video recording on a set of pre-defined neurological tests. DNE Recorder is an iOS mobile application.
It includes detailed instructions on how to perform each test alongside automated video capturing functions. Our software facilitates recording of high quality depth maps on devices equipped with LiDAR. We collect $1080 \times 720$ high-quality RGB, depth videos (upon applicable hardware) and camera calibration parameters at $60$ frames per second (FPS).
All recordings are synchronized into a secure cloud storage for offline processing. The user interface of this module is shown in Fig.~\ref{fig:dne-recorder}. 

\noindent \textbf{DNE Analyzer:} We analyze the RGB recordings offline in a separate module. The main components of DNE Analyzer include 1) vision-based pose estimation, 2) feature extraction, 3) abnormality detection. Section~\ref{sec:vision_analysis} is devoted to an elaborate description of this module.

\noindent \textbf{DNE Viewer:} Lastly, we provide a secure web application for clinicians, neurologists and researchers to monitor raw recordings view the analysis results from all subjects remotely. Fig.~\ref{fig:dne-viewer} displays a screenshot of the DNE Viewer user interface.

\vspace{-10pt}
\section{Dataset Collection}
\vspace{-5pt}
\label{sec:dataset_collection}

Our dataset collection protocol is IRB approved (\#IRB.1452500) on 02/27/2020 by the University of Illinois College of Medicine at Peoria Institute Review Board 1. In this study, 21 healthy volunteers (18 females/3 males) were recruited by sampling of convenience at the OSF HealthCare Illinois Neurological Institute Outpatient Neurology Clinic (Peoria, IL). 
Neurological examinations examine fine motor and mobility abilities. We study the FT, FTF, FR for fine motor tasks, and evaluate the mobility by the SAW test. Below we describe in detail how these tasks are performed.

\begin{itemize}
    \item \textbf{FT:} Participants are instructed to put their hands within the camera view when their index fingers and thumbs were touched. Then they would start tapping them as big open and close, and fast as they could for $15$ seconds.
    \item  \textbf{FR:} Participants are asked to gently clench their hands, hold their forearms horizontally, and roll their hands around each other as fast as possible for $15$ seconds.
    \item  \textbf{FTF:} Participants repetitively first point their index fingers towards the ceiling and then touch their fingers together out in front of their chests for a duration of $15$ seconds.
    \item \textbf{SAW:} Participants stand from a sitting pose in a chair, move the chair out of the way, walk back and forth $15$ feet. The designated time for SAW test is $45$ seconds.
\end{itemize}
 
\begin{figure}[t!]
    \centering
    \includegraphics[width=.88\linewidth]{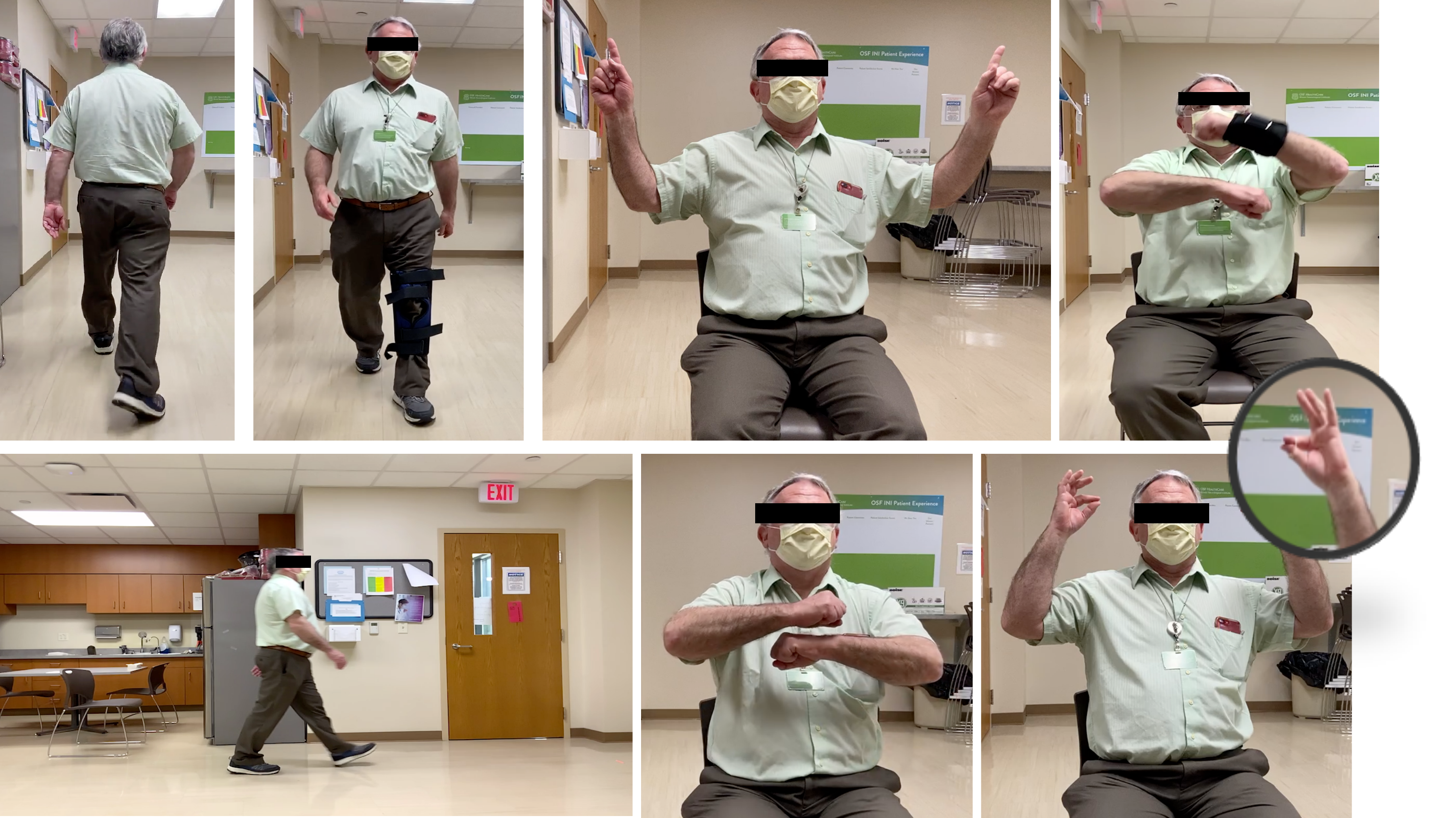}
    \vspace{-5pt}
    \caption{Examples of DNE dataset recordings. Impairments are induced by wearing a wrist brace for FR, a rubber band for FT and a knee brace for SAW tests.}
    \label{fig:dne_dataset}
\end{figure}

Each subject took two sets of neurological examinations supervised by a neurologist. In the first set of examinations, the subjects performed the tasks normally. However, for the second set, the subjects were asked to simulate motor dysfunction, i.e. perform the test abnormally. For this purpose, the subjects wore devices to deliberately add disruption to their performance and mimic impairments. For FT, a rubber band is used to restrict movements of the index and thumb fingers. For the FR and SAW tests the subjects put on a left wrist and a knee brace, respectively. On the other hand, for the FTF test, the subjects were asked to deliberately mimic a tremor pattern in moving their fingers and hands. Snapshots of recordings and subjects wearing the devices are exhibited in Fig.~\ref{fig:dne_dataset}. 

Both set of recordings are acquired by our DNE Recorder on iPad 11 Pro and iPhone 11 devices. For upper body tests, we have a close-up frontal view of the subjects with visible pelvis. Moreover, to assess the invariance of our analysis under small deviations from the frontal camera view, the view of the recordings taken on iPhone is slightly to the left compared to the iPad recordings. In addition, for the SAW, we record both saggital and frontal views, using iPad and iPhone, respectively. In total, including all four tests (FR, FT, FTF, SAW), we collect $375$ videos. Table \ref{tab:dne-dataset-statistic} provides a summary of our dataset. %

\begin{table}[]
\centering
\caption{Summary of our DNE dataset.
}
\vspace{-5pt}
\label{tab:dne-dataset-statistic}
\resizebox{.9\columnwidth}{!}{\begin{tabular}{@{}cccccccc@{}}
\toprule
\multirow{2}{*}{\textbf{Test}} &
  \multirow{2}{*}{\textbf{Total}} &
  \multicolumn{2}{c}{\textbf{Label}} &
  \multicolumn{2}{c}{\textbf{View}} &
  \multicolumn{2}{c}{\textbf{Video}} \\ 
  \cmidrule(lr){3-4}\cmidrule(lr){5-6}\cmidrule(lr){7-8}
  &&\textbf{Normal} &
  \textbf{Abnormal} &
  \textbf{Front} &
  \textbf{Side} &
  \textbf{RGB/D} &
  \textbf{RGB} \\ \midrule
FT  & 95  & 41 & 54 & 95 & -  & 45 & 50 \\
FR  & 92  & 47 & 45 & 92 & -  & 40 & 52 \\
FTF & 85  & 41 & 44 & 85 & -  & 45 & 40 \\
SAW & 103 & 41 & 62 & 61 & 42 & 54 & 49 \\ \bottomrule
\end{tabular}}
\end{table}

While there is hardly any similar publicly available upper-body neurological related dataset, there are several datasets studying gait impairments specifically in~\cite{Xue2018, sato_quantifying_2019, ortells_vision-based_2018, nieto-hidalgo_gait_2018, kidzinski_deep_2020}. The closest to our dataset is KIMORE~\cite{kimore} focusing on rehabilitation exercises rather than neurological tests. The KIMORE provides RGB, depth, and pose data for each recording, collected by Kinect v2 which is not as ubiquitous as handheld devices adopted in DNE. 
In Table \ref{tab:gait-datasets-compare}, we compare our dataset versus state of the art public gait impairment datasets in various aspects. For this comparison, we only focus on studies using a single-view, portable camera for data collection, similar to our setting. Accordingly, we list the contributions introduced by our dataset as: 1) This is the first public dataset studying multiple neurological test segments. 2) Our dataset includes normal and abnormal performance of the same task for each particular subject.
3) Our dataset contains multiple data modalities, including depth videos, camera parameters, and 2D/3D pose estimation. 

\begin{table*}[t]
\centering
\caption{A comparison between multiple vision-based gait impairment video datasets, acquired by a single camera.}
\vspace{-5pt}
\label{tab:gait-datasets-compare}
\resizebox{.95\linewidth}{!}{
\begin{tabular}{@{}cccccccccc@{}}
\toprule
\textbf{Dataset} &
  \textbf{Availability} &
  \textbf{\begin{tabular}[c]{@{}c@{}}Sagittal\\ View\end{tabular}} &
  \textbf{\begin{tabular}[c]{@{}c@{}}Frontal\\ View\end{tabular}} &
  \textbf{\begin{tabular}[c]{@{}c@{}}Data\\ Type\end{tabular}} &
  \textbf{\begin{tabular}[c]{@{}c@{}}Mobile \\ Device\end{tabular}} &
  \textbf{\begin{tabular}[c]{@{}c@{}}Number of \\ Subjects\end{tabular}} &
  \textbf{\begin{tabular}[c]{@{}c@{}}Number of\\ Sequences\end{tabular}} &
  \textbf{\begin{tabular}[c]{@{}c@{}}Pose \\ Estimation\end{tabular}} &
  \textbf{\begin{tabular}[c]{@{}c@{}}Normal and \\ Abnormal Pairs\end{tabular}} \\ \midrule
Xue et al. \cite{Xue2018} & \xmark & -      & -      & RGB    & \xmark & -    & -    & 2D     & \xmark \\ 
Sato et al. \cite{sato_quantifying_2019}            & \xmark & \xmark & \cmark & RGB    & \xmark & 2    & 2    & 2D     & \xmark \\ 
Ortells et al. \cite{ortells_vision-based_2018}     & \cmark & \xmark & \cmark & Binary & \xmark & 10   & 20   & \xmark & \cmark \\
Nieto-Hidalgo et al. \cite{nieto-hidalgo_gait_2018} & \cmark & \cmark & \cmark & Binary & \cmark & -    & 73   & \xmark & \cmark \\ 
Kidzinski et al. \cite{kidzinski_deep_2020}         & \cmark & \cmark & \xmark & RGB & \xmark & 1026 & 1792 & 2D     & \xmark \\
Ours                                            & \cmark & \cmark & \cmark & RGB/D  & \cmark & 21   & 336  & 2D/3D  & \cmark \\ \bottomrule
\end{tabular}
}
\vspace{-6mm}
\end{table*}

\vspace{-8pt}
\section{DNE Vision-Based Analysis}
\label{sec:vision_analysis}
\begin{figure}[!t]
\centerline{\includegraphics[width=0.85\columnwidth]{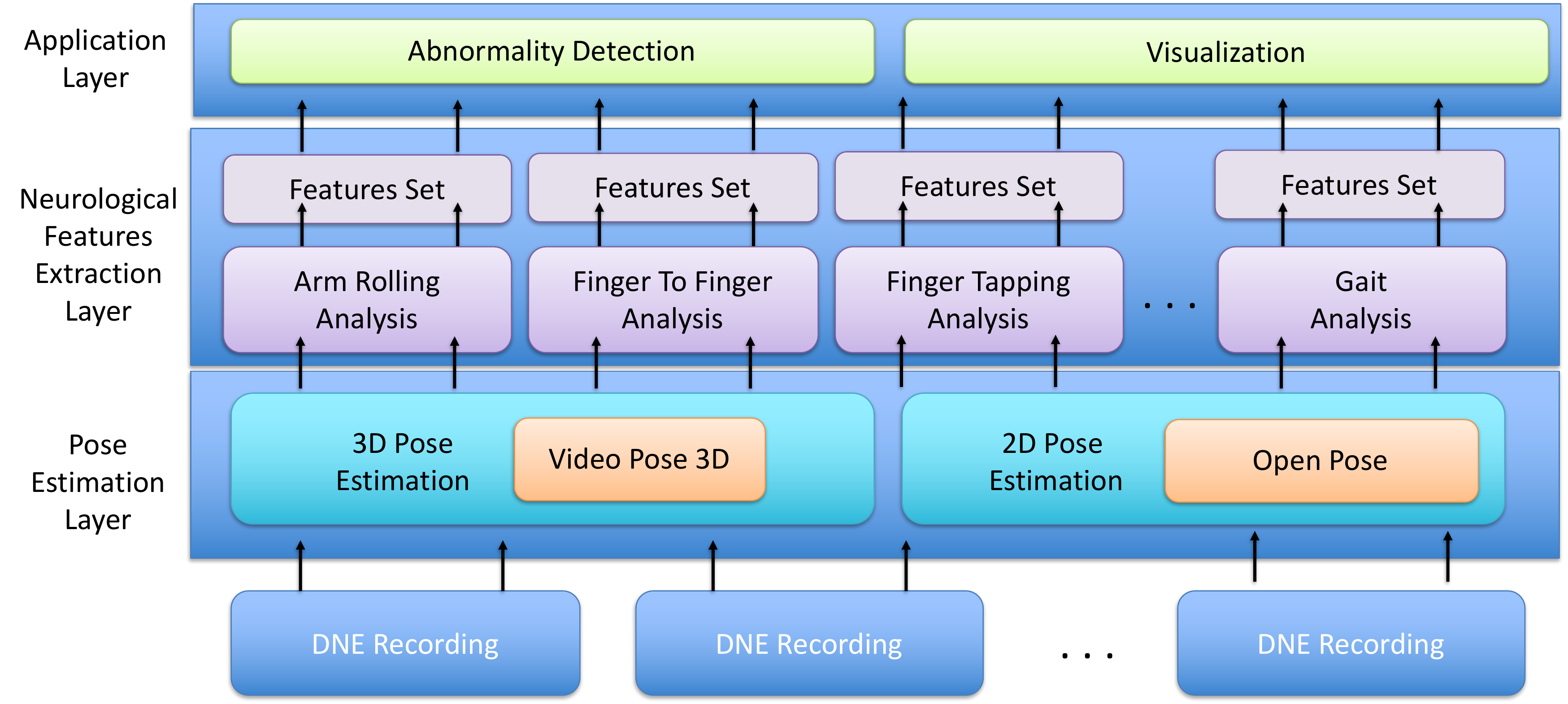}}
\vspace{-5pt}
\caption{Overview of DNE vision-based analysis framework.} 
\label{fig:overall_framework}
\end{figure}
In our DNE analysis pipeline, given an RGB video, we first compute the human pose in each frame. Next, from the pose time series, we extract a set of features that quantify the subject's performance in various aspects.
We structure our analysis pipeline into three layers, namely 1) pose estimation, 2) feature extraction, and 3) application layer, as illustrated in Fig.~\ref{fig:overall_framework}.
The pose estimation layer provides frame-level high-quality 2D/3D joint locations (Section \ref{ssec:pose-estimation}).
We pre-process the estimated pose to prepare it for feature computation. In the feature extraction layer, we calculate a set of features that describe subject's performance on various tests. We carefully design these features for each test separately to accurately reflect the subjects performance and dedicated abnormalities.
Lastly, the application layer contains several downstream tasks consuming the features, including abnormality detection and visualization for a qualitative comparison among recordings.

\vspace{-10pt}
\subsection{Pose estimation}
\vspace{-2pt}
\label{ssec:pose-estimation}
For upper body tests (FT, FTF, and FR), we use OpenPose (OP)~\cite{Cao2019_openpose} to estimate the 2D pose. On the other hand, for SAW tests, we compute the 3D pose using the VideoPose3D (VP3D) package~\cite{pavllo2019_videopose3d}. 
OP is designed for multi-person real time 2D hand \cite{Simon2017} and body ~\cite{Cao2019_openpose} pose estimation. Given an RGB image, OP first detects all visible body parts in the image and the corresponding part affinity fields. Associating body parts to each individual in an image boils down to a graph matching problem which can be solved in a greedy fashion. 
Meanwhile, VP3D adopts dilated temporal convolution to estimates 3D pose from sequence of 2D keypoints extracted from the video.

For upper body tests, if the subject and the moving limb is located parallel to the camera plane, then the motion is well approximated in a plane, i.e. in two dimensions. That is why 2D pose is chosen for upper body tests. However, this might not hold for the SAW test (especially depending on the camera view), hence urging us to use 3D pose for this analysis. 

We use OP and VP3D to extract 2D/3D pose of the recordings in the DNE dataset. We measure the inference time of pre-trained OP models on Google Colab with one Tesla T4 GPU as $6$ FPS. In VP3D, the 2D pose data are extracted for each frame via Detectron2~\cite{wu2019detectron2}, which achieves the  runtime of $12.5$ FPS on a GeForce RTX 2080 GPU. Additionally, it takes roughly $0.2$ seconds for uplifting 2D pose to 3D per video.
\vspace{-10pt}
\subsection{Pre-processing}
\vspace{-2pt}
\label{sec:prep}
We truncate a recording to only include the sequence of frames that are related to the subject performing the test. 
To account for variable distance of the subjects from the camera, we normalize the estimated pose by a reference length. For FT, FTF and FR tests, the reference is the length of the forearm. For SAW, the reference is the distance between the pelvis and neck joints. We compute the reference lengths as the median of the value across all the frames. 
In addition, as the estimated pose can be erroneous at some frames we use median and Savitzky-Golay filtering~\cite{Savitzky1964}. In our dataset, we have excluded $27$ recordings due to unreliable and noisy estimated pose. 
Therefore, we only analyzed $348$ videos in total. %

\vspace{-10pt}
\subsection{Notations}
\vspace{-2pt}

\begin{figure}
    \centering
    \includegraphics[width=.8 \linewidth]{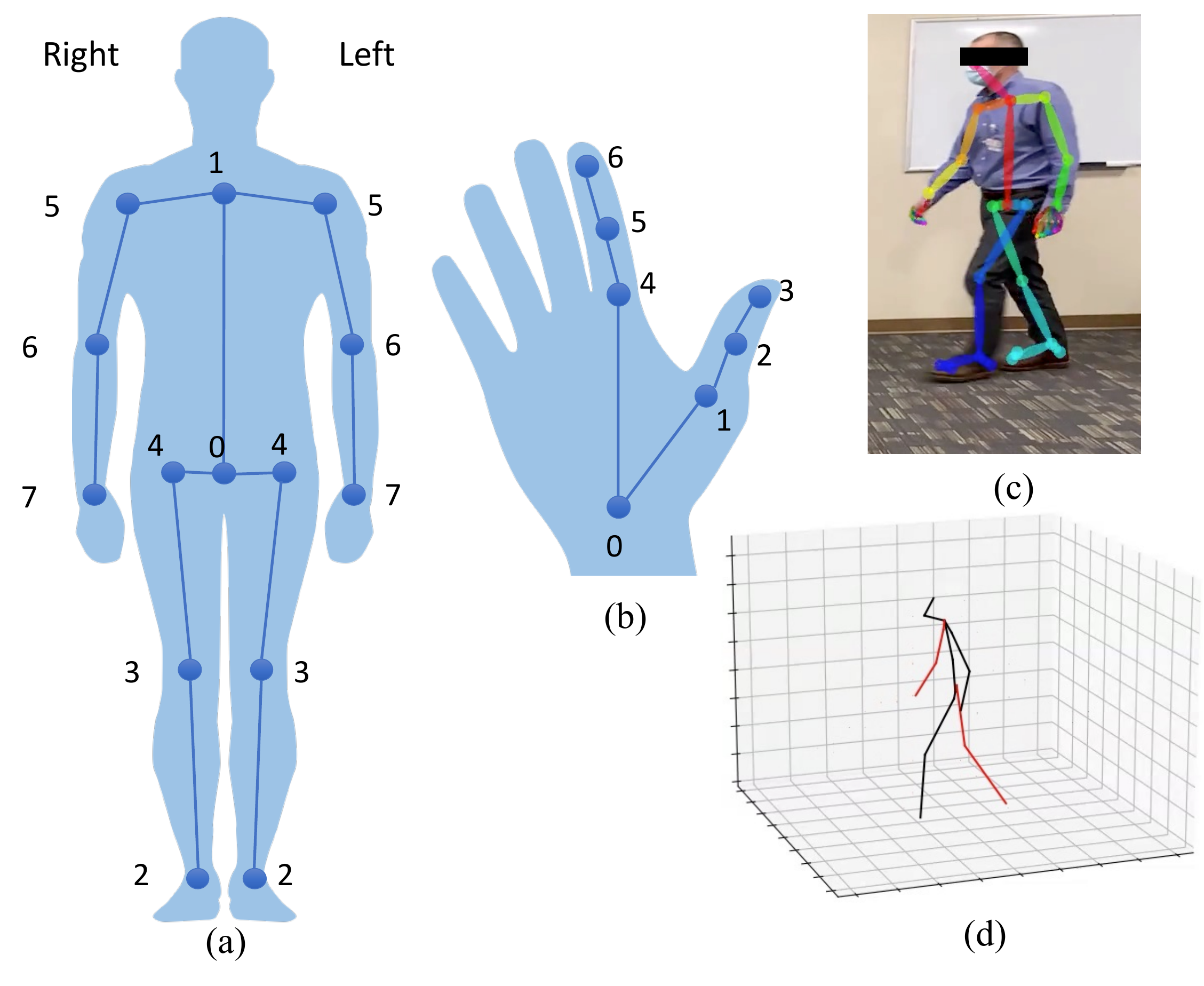}
    \vspace{-5pt}
    \caption{Skeleton tree for (a) body ${{B}}$, and (b) hand ${{H}}$. Examples of human pose estimation (c) in 2D (${B}_2, {H}_2$) using OpenPose~\cite{Cao2019_openpose} and (d) in 3D (${B}_3$) using VideoPose3D~\cite{pavllo2019_videopose3d}.} 
    \label{fig:pose_estimation}
\end{figure}
Given the pose sequence estimated from the RGB video, we extract a set of quantified features. Below, we first express our notations and then introduce the features we defined for each test. Let $\boldsymbol{v} = [v_1, ..., v_N]$ denote the set of $N$ frames ordered chronologically in video $\boldsymbol{v}$. There is a one-to-one correspondence between the time associated with each frame and the frame index,  where $\boldsymbol{t} = [t_1, ..., t_N]$ and $t_i = i/fps$, $fps$ denoting the frame per second rate of the video. Given $\boldsymbol{v}$ and the pose estimation module (such as OP or VP3D), we extract the location of $K$ keypoints in each frame. For convenience, we use the same indexing of the body joints for both 2D and 3D pose. However, to differentiate between the 2D and 3D pose, we denote each by ${B}_2$ and ${B}_3$, respectively. Furthermore, we use ${H}_2$ to represent the 2D hand keypoints. An illustration of the hand and body skeleton trees alongside our indexing notations are provided in Fig.~\ref{fig:pose_estimation}. Note that, for the sake of brevity, we have only indexed a subset of the keypoints that we are using in our analysis.

We reserve $\boldsymbol{s}_{k, *}[i]$ for the location of the $k$-th keypoint at frame $i$, corresponding to skeleton tree $* \in \{{H}_2, {B}_2, {B}_3\}$. For $* \in \{{H}_2, {B}_2\}$, $\boldsymbol{s}_{k, *}[i] \in \mathbb{R}^2$ and for $*={B}_3$, $\boldsymbol{s}_{k, *}[i] \in \mathbb{R}^3$. Furthermore, we add superscript $r$ and $l$ to point to right and left (R/L) body parts, respectively. For example, $\boldsymbol{s}^r_{3, {H}_2}[i]$ locates the tip of the right thumb at frame $i$.

To extract kinematic features that quantify the performance of a subject in a test, we track the location of various major keypoints and define a set of features accordingly. Major keypoints vary based on the test. For instance, the major keypoints in FT include the tip of the index and thumb fingers of two hands while in FR, we closely track the wrist joints. 

In different tests, the subjects are asked to move certain limbs repeatedly. 
Thus, it is natural to compute features such as frequency, and amplitude for periodic pose patterns and report the mean and standard deviations (STD) across different cycles. In addition, for a test performed normally, the features 
corresponding to the R/L body parts should be close. Thus, to quantify the difference between the right ${f}^r$ and left ${f}^l$ features, we define an asymmetry metric as:
\begin{align}
   \textrm{Asym}({f}^r, {f}^l) = \frac{\vert {f}^r- {f}^l \vert}{{f}^r + {f}^l}. \label{eq:asym} 
\end{align}
Another useful metric in our analysis is Pearson correlation coefficient denoted by $\text{CC}$. For two 1D discrete time series $\boldsymbol{x}_1$ and $\boldsymbol{x}_2$, we define $\text{CC}$ as:
\vspace{-5pt}
\begin{align}
    \textrm{CC}(\boldsymbol{x}_1, \boldsymbol{x}_2) = \frac{(\boldsymbol{x}_1-\bar{\boldsymbol{x}}_1)^T (\boldsymbol{x}_2-\bar{\boldsymbol{x}}_2)}{\Vert \boldsymbol{x}_1-\bar{\boldsymbol{x}}_1 \Vert_2 \Vert \boldsymbol{x}_2-\bar{\boldsymbol{x}}_2 \Vert_2}.\label{eq:cc}
\end{align}
where $\bar{.}$ and $.^T$ denote the mean and transpose operators.
For highly correlated series, $\vert \textrm{CC} \vert$ is close to one.

\vspace{-8pt}
\subsection{Feature definition}
\vspace{-2pt}
We list the features defined for various tests in Table~\ref{tab:dne-features} and describe them in detail below.

\begin{figure}%
\pgfplotsset{every x tick label/.append style={font=\tiny, yshift=0.5ex}}
\pgfplotsset{every y tick label/.append style={font=\tiny, xshift=0.5ex}}
\centering

    \includegraphics[]{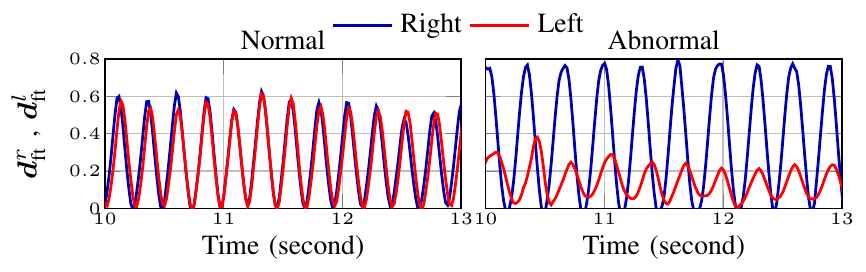}
\vspace{-0.7cm}
\caption{FT amplitude for normal and abnormal examples.} 
\label{fig:ft_features}
\end{figure}

\noindent \textbf{Finger Tapping (FT)}: For this test, the major keypoints are the tip of the R/L thumb and index fingers alongside R/L wrist and elbow joints. To extract properties of the periodic motion, we examine the distance between the tip of the index and thumb fingers across time defined as:

\begin{align}
    \boldsymbol{d}_{\text{ft}}^*[i] = \Vert \boldsymbol{s}^*_{3, {H}_2}[i] - \boldsymbol{s}^*_{6, {H}_2}[i] \Vert_2, \quad * \in \{r, l\}. \label{eq:dist_ft}
\end{align}
Examples of $\boldsymbol{d}_{\text{ft}}^r$ and $\boldsymbol{d}_{\text{ft}}^l$ for normal and abnormal executions of the FT test are provided in Fig.~\ref{fig:ft_features}. In our dataset, to simulate abnormality in FT the subjects are wearing a rubber band around index and thumb fingers of one hand. As also revealed in Fig.~\ref{fig:ft_features}, this limits the tapping amplitude of the hand wearing the band and slows down the tapping rate.
Given $\boldsymbol{d}_{\text{ft}}^*$, we compute the period for the $*$ hand, $T_{\textrm{ft}}^*$, as the time (in seconds) between two consecutive local minima (or maxima) of $\boldsymbol{d}_{\text{ft}}^{*}$. Frequency $F_{\textrm{ft}}^*$ is the reciprocal of $T_{\textrm{ft}}^*$. 
We also report the magnitude of finger-tapping $A^*_{\textrm{ft}}$ as the difference in consecutive minima and maxima of $\boldsymbol{d}_{\text{ft}}^*$. We also report the asymmetry of the periods ($\text{Asym}(T_{\text{ft}}^{r}, T_{\text{ft}}^{l})$), frequencies ($\text{Asym}(F_{\text{ft}}^{r}, F_{\text{ft}}^{l})$) 
and amplitudes ($\text{Asym}(A_{\text{ft}}^{r}, A_{\text{ft}}^{l})$) of R/L hands following~\eqref{eq:asym}.

\begin{figure}[t!]
    \pgfplotsset{every x tick label/.append style={font=\tiny, yshift=1.5ex}}
    \pgfplotsset{every y tick label/.append style={font=\tiny, xshift=1.5ex}}
    \centering
    \begin{subfigure}[b]{\columnwidth}
    \centering
    \includegraphics[]{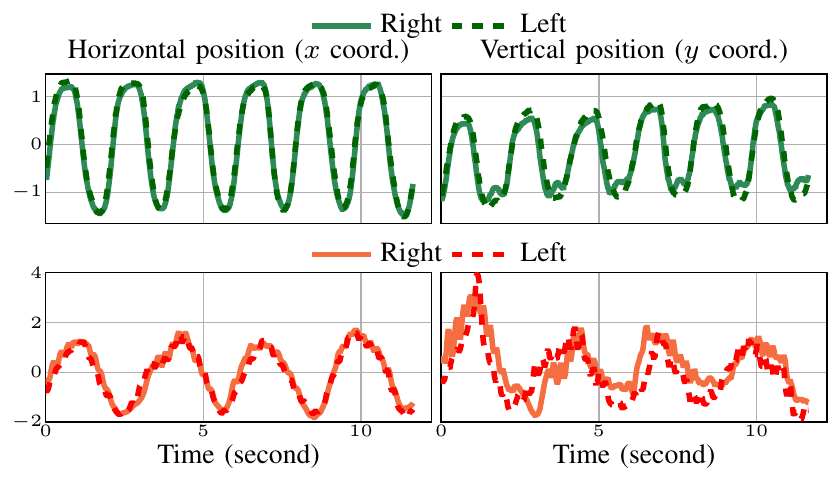}
    \vspace{-5pt}
    \caption{Horizontal and vertical (first/second column) locations of joint index $5$ in ${H}_2$ for normal and abnormal (first/second row) recordings.
    }
    \label{fig:ftf-lr-asymmetry}
    \end{subfigure}
    \par \medskip
    \begin{subfigure}[b]{\columnwidth}
    \centering
    \includegraphics[]{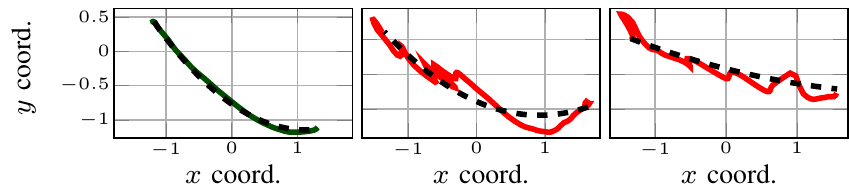}
    \vspace{-5pt}  
    \caption{Actual trajectory (solid) vs. smoothed fitted trajectory (dashed). 
    }
    \label{fig:ftf-traj}
    \end{subfigure}
    \par\medskip    
    \begin{subfigure}[b]{\columnwidth}
    \centering
    \includegraphics[width=\linewidth]{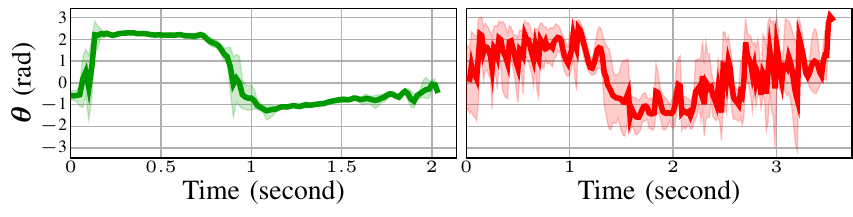}
    \vspace{-20pt}
    \caption{Cycle-wise average of velocity angle. 
    }
    \label{fig:ftf-vel-angle}
    \end{subfigure}
    \caption{FTF features including finger (a) positions, (b) spatial trajectory, (c) velocity angle. \textcolor{forestgreen}{Green} (\textcolor{red}{red}) curves stand for normal (abnormal) recordings. In each row (column), the subplots share the same vertical (horizontal) axis.}
    \label{fig:ftf-features}
\end{figure}

Furthermore, we define the instant tapping speed and acceleration for R/L hands as the first and second order derivatives of $\boldsymbol{d}_{\text{ft}}^r$ and $\boldsymbol{d}_{\text{ft}}^l$ with respect to time. We adopt mean and maximum of instant speed and acceleration across tapping cycles as features.
We also introduce average tapping rate as the average number of finger taps per second, obtained by dividing the total number of finger tappings by the FT duration.

Finally, to evaluate the stability of the hands and arms during the FT recording, we examine the wrist and elbow joints. For this purpose, we introduce the relative height between $(\boldsymbol{s}^r_{7, {B}_2},\boldsymbol{s}^l_{7, {B}_2})$ and $(\boldsymbol{s}^r_{6, {B}_2},\boldsymbol{s}^l_{6, {B}_2})$ across $N$ frames:
\vspace{-5pt}
\begin{align}
    C_{\textrm{ft}}^{\textrm{wrist}} &= \frac{1}{N} \sum\limits_{i=1}^N \frac{\Vert \boldsymbol{s}^r_{7, {B}_2}[i]-\boldsymbol{s}^l_{7, {B}_2}[i] \Vert_2}{\Vert \boldsymbol{s}^r_{7, \textrm{B}_2}[i] \Vert_2},
\end{align}
\vspace{-10pt}
\begin{align}
    C_{\textrm{ft}}^{\textrm{elbow}} &= \frac{1}{N} \sum\limits_{i=1}^N \frac{\Vert \boldsymbol{s}^r_{6, {B}_2}[i]-\boldsymbol{s}^l_{6, {B}_2}[i] \Vert_2}{\Vert \boldsymbol{s}^r_{6, {B}_2}[i] \Vert_2}. 
    \label{eq:wrist_stable}
\end{align}
\vspace{-10pt}

\begin{table*}[]
\centering
\caption{Summary of our DNE features. Asymmetry between R/L features is computed based on~\eqref{eq:asym}.}
\label{tab:dne-features}
\resizebox{.95\linewidth}{!}{
\begin{tabular}{p{.25\linewidth}p{.25\linewidth}p{.25\linewidth}p{.25\linewidth}}
\toprule
\textbf{Finger Tapping (FT)} & \textbf{Finger to Finger (FTF)} & \textbf{Forearm Roll (FR)} & \textbf{Stand and Walk (SAW)}\\
\midrule
\textbf{Amplitude (R/L)}  \par \textit{Mean, STD, Median, Asymmetry} \par Maximum distance  between the tip of the index and thumb fingers \par
\textbf{Period (R/L)}  \par \textit{Mean, STD, Median, Asymmetry} \par Time (in seconds) taken to complete one tapping cycle for R/L hands \par
\textbf{Frequency (R/L)}
\par \textit{Mean, STD, Median} \par Reciprocal of period (1/second) for R/L hands \par
\textbf{Maximum speed (R/L)}  \par \textit{Mean, Asymmetry} \par Maximum of instant tapping speed (defined as the derivative of the distance between the tip of the index and thumb fingers with respect to time) for R/L hands \par
\textbf{Maximum acceleration (R/L)} \par \textit{Mean, STD, Median, Asymmetry} \par Maximum of instant tapping acceleration (defined as the second derivative of the distance between the tip of the index and thumb fingers with respect to time) for R/L hands \par
\textbf{Average tapping rate (R/L)}
\par Total number of finger taps divided by the duration of FT test in seconds for R/L hands \par
\textbf{Wrist stability} \par \textit{Mean, STD, Median} \par Variations in R/L wrist joint positions\par
&
\textbf{Horizontal symmetry} \par \textit{CC}   \par The CC of horizontal spatial trajectory of the R/L index finger \par
\textbf{Vertical symmetry} \par \textit{CC}     \par The CC of the vertical spatial trajectory of the R/L index finger \par
\textbf{Period (R/L)}           \par \textit{Mean, STD} \par Total time (in seconds) taken for one complete cycle (moving from the highest to the lowest vertical position and back) on each side \par
\textbf{Average speed R/L}          \par \textit{Mean, STD} \par The traversed distance of R/L index fingers within half a cycle’s period divided by half the cycle’s period  \par
\textbf{Path smoothness (R/L)}  \par \textit{Mean, STD} \par The ratio between the actual traversed distance of R/L index fingers and the length of the fitted smooth curve \par
\textbf{Velocity angle symmetry (R/L)} \par \textit{Mean, STD} \par The pairwise CC between the angle velocity series of any two cycles \par

&
\textbf{Amplitude R/L}  \par \textit{Mean, STD, Median, Asymmetry} \par Distance between the minimum and maximimum of the vertical position of the R/L wrists \par
\textbf{Period R/L}  \par \textit{Mean, STD, Median, Asymmetry} \par Time (in seconds) taken to complete one forearm roll cycle for R/L hands \par
\textbf{Maximum speed (R/L)} \par \textit{Mean, STD, Median, Asymmetry} \par Maximum of forearm roll speed (defined as the first derivative of the vertical coordinate of the wrist joint with respect to time) for R/L hands \par
\textbf{Maximum acceleration (R/L)} \par \textit{Mean, STD, Median, Asymmetry} \par Maximum of forearm roll acceleration (defined as the second derivative of the vertical coordinate of the wrist joint with respect to time) for R/L hands \par
\textbf{Rolling speed R/L}\par \textit{Mean, STD, Median} \par Average forearm roll speed (defined as the amplitude divided by half the rolling cycle period) \par
\textbf{Average rolling rate R/L}
\par Total number of forearm roll cycles divided by the duration of FR test in seconds for R/L hands \par
&
\textbf{Knee angle symmetry}    \par \textit{Mean, STD, Median} \par The CC of the aligned R/L knee angle series within a walking segment (a full pass of the room length) \par
\textbf{Step symmetry}  \par \textit{Mean, STD, Median} \par The cycle-wise CC of the aligned spatial trajectory of the R/L foot in the horizontal axis  \par
\textbf{Step length}    \par \textit{Mean, STD, Median} \par The furthest distance between two feet within each step \par
\textbf{Step width}    \par \textit{Mean, STD, Median} \par The shortest distance between two feet within each step \par
\textbf{Step time}    \par \textit{Mean, STD, Median} \par The time (in seconds) to complete one step (the interval between two consecutive time-points having the shortest distance between two feet) \par
\textbf{Time to stand}   \par Total time taken (in seconds) from the first stand up effort to a full standing on feet state \par
\textbf{Turning time}    \par \textit{Mean, STD, Median} \par Total time taken (in seconds) for a subject to turn around after each walking segment \par
\textbf{Walking speed}    \par \textit{Mean, STD} \par Total of traveled distance of the pelvis joint divided by the duration of a walking segment \par
\textbf{Cadence}    \par \textit{Mean, STD} \par Total number of steps divided by the duration of a walking segment \par
\\
\bottomrule
\end{tabular}}
\vspace{-4mm}
\end{table*}

\noindent \textbf{Finger to Finger (FTF)}: 
In our dataset, we observe that the estimated pose by OP for middle joints of the index finger, i.e. joint index $5$ in ${H}_2$, is more stable than the outer fingertip. Hence, we focus on this joint for FTF test.
In a normal FTF, the horizontal and vertical trajectories of the R/L hands are symmetric up to a mirroring (Fig.~\ref{fig:ftf-lr-asymmetry}- top row), while this does not necessarily hold for abnormal case (Fig.~\ref{fig:ftf-lr-asymmetry}- bottom row). Thus, in each cycle, we define the cross correlation of the R/L horizontal ($x$) and vertical ($y$) coordinates as the horizontal ($S_{\text{ftf}}^{\text{finger-x}}$) and vertical symmetries ($S_{\text{ftf}}^{\text{finger-y}}$):

\begin{align}
    \label{eq:ftf_finger}
   S_{\text{ftf}}^{\text{finger-x}} &= \text{CC}(\left[\boldsymbol{s}_{5, {H}_2}^l\right]_x, - \left[\boldsymbol{s}_{5, {H}_2}^r\right]_x), \\
   S_{\text{ftf}}^{\text{finger-y}} &= \text{CC}(\left[\boldsymbol{s}_{5, {H}_2}^l\right]_y, \left[\boldsymbol{s}_{5, {H}_2}^r\right]_y)
\end{align}
\noindent where $\left[\boldsymbol{s}_{5, H_2}^* \right]_\dagger = \{\boldsymbol{s}^*_{5, {H}_2}[i](\dagger)\}_{i=1}^{N}$, $\dagger \in \{x, y\}$ and $* \in \{r, l\}$, is the $x$ or $y$ coordinates of the pose series.
We also compute the period and average speed. We derive the average speed by dividing the traversed distance of R/L finger within half a cycle's period by half the cycle's period.

Patients with neurological impairments tend to have tremors while moving their fingers during FTF test \cite{krishna_quantitative_2019}. This leads to a deviation of the fingers' trajectory from a smooth curve. To characterize this deviation, we first fit a smooth curve to the fingers' trajectory, in the form of a second order polynomial in terms of the $x$ and $y$ coordinates. We observe that fitting a second order function to the trajectories, well matches the FTF trajectories of normal subjects. We consider the length of this smooth curve as a reference to compare against the length of the original fingers' trajectory. We then define the ratio of the length of the actual fingers' trajectory during each FTF cycle by the length of the fitted smooth curve as path smoothness metric (PS). We report PS for R/L hands. Examples of normal and abnormal finger trajectories alongside the smooth fitted curves are plotted in Fig.~\ref{fig:ftf-traj}. %

Another feature we found helpful in detecting abnormal function in FTF is instant velocity. We derive the instant velocity vector by the first derivative of the horizontal and vertical pose with respect to time.
We then examine the angle between the vertical and horizontal components of this vector on the R/L hands. At time instant $t$, the \textit{velocity angle} $\boldsymbol{\theta}$ is:
\begin{align}
    \boldsymbol{\theta}^{*}(t) = \textrm{atan2} \left(\frac{\frac{d \, \left[\boldsymbol{s}^*_{5, {H}_2}\right]_{y}}{d t}}{\frac{d \, \left[\boldsymbol{s}^*_{5, {H}_2}\right]_x}{d t}} \right), * \in \{r, l\}. \label{eq:vel_ang}
\end{align}
Next, for each hand, we compare $\boldsymbol{\theta}$ across different cycles using CC in~\eqref{eq:cc}. Given $N_C$ number of cycles, we have $N_C \choose 2$ CC values assessing the symmetry of the R/L velocity angles across different cycles, which we summarize by reporting the mean and STD. Examples of normal and abnormal aligned velocity angles across different cycles are provided in Fig.~\ref{fig:ftf-vel-angle}. Note that for abnormal FTF, large magnitude fluctuations, caused by tremors in moving the hands, visibly appear in $\boldsymbol{\theta}$.

\begin{figure}%
\pgfplotsset{every x tick label/.append style={font=\tiny, yshift=0.5ex}}
\pgfplotsset{every y tick label/.append style={font=\tiny, xshift=0.5ex}}
\centering

\includegraphics[]{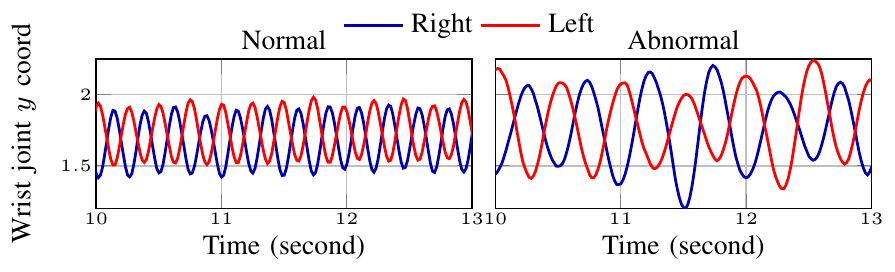}
\caption{Vertical ($y$) coordinate of the wrist joint versus time for normal and abnormal examples in FR test.} 
\label{fig:fr_example}
\end{figure}

\noindent \textbf{Forearm Rolling (FR)}: We include the wrist and elbow joints as the major keypoints for this test. We specifically attend to the vertical coordinate of the wrist joints to compute period $T^*_\textrm{fr}$
and amplitudes $A^*_\textrm{fr}$ for $* \in \{r, l\}$. Fig.~\ref{fig:fr_example} illustrates the vertical position of the R/L wrists for a normal and abnormal example. Note that, due to wearing the device in the abnormal recording, the period of the forearm roll cycles for both R/L hands are larger compared to its normal counterpart. In addition, similar to FT, we include the asymmetry of the aforementioned metrics in the FR features.

We also include the maximum instant speed and acceleration derived from vertical coordinates of the wrist joints. Similar to FT, we define rolling speed and rate. Rolling speed is computed as the difference between the minimum and maximum of $y$ coordinate of the R/L hands divided by half the rolling period. Also, rolling rate is defined as the number of rolling cycles per second. Finally, we report the stability of the elbows $C_{\textrm{fr}}^{\textrm{elbow}}$ and define it analogous to~\eqref{eq:wrist_stable}. %

\begin{figure}[t!]
    \centering
    \begin{subfigure}[b]{\columnwidth}
         \centering
        \includegraphics[]{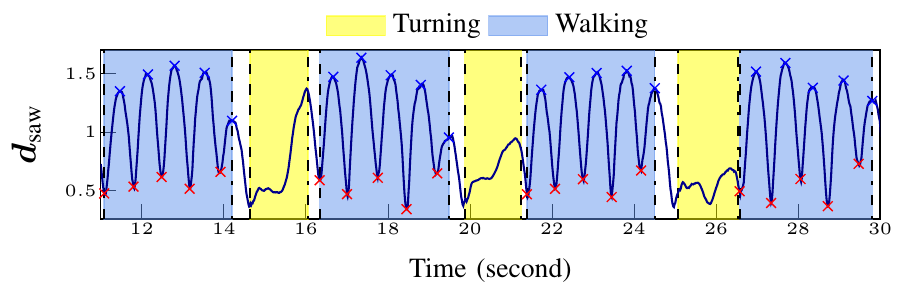}
    \vspace{-15pt}
    \caption{Distance between two feet ($d_{\textrm{saw}}$) versus time. Marker \textcolor{red}{{\ding{53}}} and \textcolor{blue}{{\ding{53}}} denote the start and end of each step. In this example, a SAW video is partitioned into 4 walking (W) and 3 turning (TU) segments.}
    \label{fig:saw-step-info}
    \end{subfigure}
    \par\medskip
    \begin{subfigure}[b]{\columnwidth}
        \centering
        \includegraphics[]{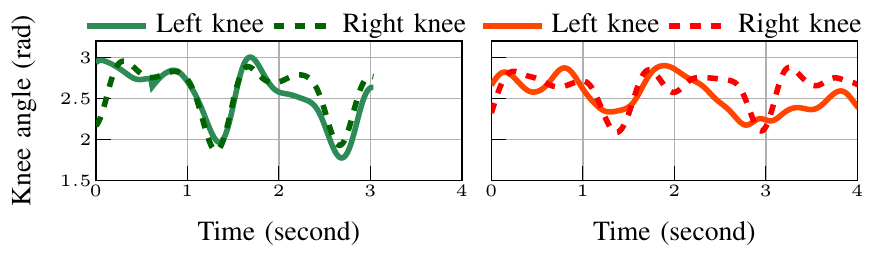}
    \vspace{-20pt}
    \caption{Knee angle series for a walking segment taken from a normal and an abnormal recording of the same subject. \textcolor{forestgreen}{Green} (\textcolor{red}{red}) curves stand for normal (abnormal) recordings.}
    \label{fig:saw-knee-angle}
    \end{subfigure}
     
     \caption{Examples of SAW features.}
\end{figure}

\noindent \textbf{Stand-up and Walk (SAW):} We use the side-view SAW recordings in our analysis of SAW test. For SAW pose estimation, we use VP3D~\cite{pavllo2019_videopose3d}. In VP3D, the joint locations are defined relative to the pelvis joint. As a result, estimated pose by VP3D misses the global position of subjects within a frame which is essential to detect different segments
of the SAW test, i.e. stand-up (SU), walk (W), and turn (TU). This urged us to track the 2D position of the pelvis $\boldsymbol{s}_{0,B_2}$ extracted by OP as a notion of subject's global position in a video frame. Analyzing this position through time enables us to split a SAW recording into multiple non-overlapping SU, W, and TU segments. Supplementary Fig. S3 visualizes these segments.

For the SU segment of SAW, we focus on the time to stand~\cite{Jawad2012}, measured by the total time taken from the first SU effort to a full standing on feet state.
We derive time to stand by thresholding the magnitude of the pelvis joint's velocity.
Note that, since our subjects are asked to walk back and forth a designated room multiple times, at some points, they have to change direction and turn around. We report time to turn around as another indicative feature for SAW test.

The first set of features derived for the walking segment are obtained based on the distance between the two feet stated as:
\vspace{-15pt}
\begin{align}
    \boldsymbol{d}_{\textrm{saw}}[i] = ||\boldsymbol{s}^r_{2,{B}_3}[i]-\boldsymbol{s}^l_{2,{B}_3}[i]||_2.
\end{align}
Note that, the periodic nature of a normal gait also reflects in $\boldsymbol{d}_{\textrm{saw}}$ (see Fig.~\ref{fig:saw-step-info}). Given $\boldsymbol{d}_{\textrm{saw}}$, we highlight different W and TU segments in Fig.~\ref{fig:saw-step-info}. For a gait pattern derived based on $\boldsymbol{d}_{\textrm{saw}}$, \textit{step time} is the time to complete one step and computed as the time difference between two consecutive local maxima of $\boldsymbol{d}_{\textrm{saw}}$. Meanwhile, \textit{step length} defined as linear distance between two successive placements of the same foot~\cite{Herran2014} manifests as the local maxima of $\boldsymbol{d}_{\textrm{saw}}$. The \textit{step width}, on the other hand, is interpreted as the local minima of $\boldsymbol{d}_{\textrm{saw}}$. The calculations of these features in turning segments are excluded.

As two global features for gait, we report mean and STD of \textit{cadence} and \textit{average speed} across all W segments. We compute cadence as the number of steps divided by the duration of a walking segment. Average speed is determined by the total traveled distance of the pelvis joint divided by the duration of a walking segment.

To evaluate the symmetry of the R/L gait, we introduce the cross correlation between the knee angle series of R/L legs, denoted by {${S}_{\textrm{saw}}^{\textrm{knee angle}}$}. We find this feature a good descriptive of gait abnormality, as in our recordings, gait abnormality is introduced through wearing a knee band which limits the knee motion (Fig.~\ref{fig:dne_dataset}). For each frame, we define the knee angle as the angle between $\boldsymbol{s}^r_{3, {B}_3}-\boldsymbol{s}^r_{4, {B}_3}$ and $\boldsymbol{s}^r_{3, {B}_3}-\boldsymbol{s}^r_{2, {B}_3}$ for the right leg and $\boldsymbol{s}^l_{3, {B}_3}-\boldsymbol{s}^l_{4, {B}_3}$ and $\boldsymbol{s}^l_{3, {B}_3}-\boldsymbol{s}^l_{2, {B}_3}$ for the left leg. As there is a lag between the R/L gait cycles, we align the knee angle series of the R/L legs within each cycle and then report CC of the aligned series. Examples of aligned normal and abnormal knee angles for R/L legs are shown in Fig.~\ref{fig:saw-knee-angle}. For normal gait, the R/L knee angles are highly correlated after alignment (Fig.~\ref{fig:saw-knee-angle}-top row), while this does not hold for abnormal gait (Fig.~\ref{fig:saw-knee-angle}-bottom row).  

In addition, we define step symmetry between the R/L feet movements by comparing the horizontal position of R/L feet at different gait cycles. We represent this metric by {${S}_{\textrm{saw}}^{\textrm{feet-x}}$}. To compute {${S}_{\textrm{saw}}^{\textrm{feet-x}}$}, similar to {${S}_{\textrm{saw}}^{\textrm{knee angle}}$}, we first align the R/L horizontal positions within each gait stride and report the CC of the aligned series. We report mean and STD for both {${S}_{\textrm{saw}}^{\textrm{feet-x}}$} and {${S}_{\textrm{saw}}^{\textrm{knee angle}}$} across different cycles. %

\vspace{-10pt}
\section{Results and Discussion}
\label{sec:results}

\subsection{Subject-based Normal vs. Abnormal Comparison}
\vspace{-2pt}
In this section, we aim to compare the normal and simulated-impaired performances of the same subject and show that this analysis is insensitive to the choice of recording device and robust to the viewpoint or distance from the camera. Note that in our dataset, for each subject, we have four sets of recordings. Two of these recordings capture the normal performance of the test, while in the other two, the subject is asked to perform abnormally. In addition, two pairs of normal/abnormal recordings are captured by an iPhone (P) and an iPad (T). Let $N_P$/$N_T$ and $A_P$/$A_T$ denote the normal and abnormal recordings captured by iPhone/iPad.

For each feature and subject, we define A-A/N-N as the intra-class distance between the features derived from the abnormal/normal recordings of the subject captured on iPhone and iPad devices. In other words, A-A is the distance between features computed for $A_T$ and $A_P$ recordings, while N-N marks the difference between the features of $N_T$ and $N_P$ videos. 
For N-A, we consider the distance between $A_T$-$N_P$ and $N_T$-$A_P$ pairs and report the average.
We normalize the A-A, N-N, and N-A distances by the maximum of N-A distances.

\begin{figure}[t!]
    \centering
    \resizebox{.92\columnwidth}{!}{
        \includegraphics[]{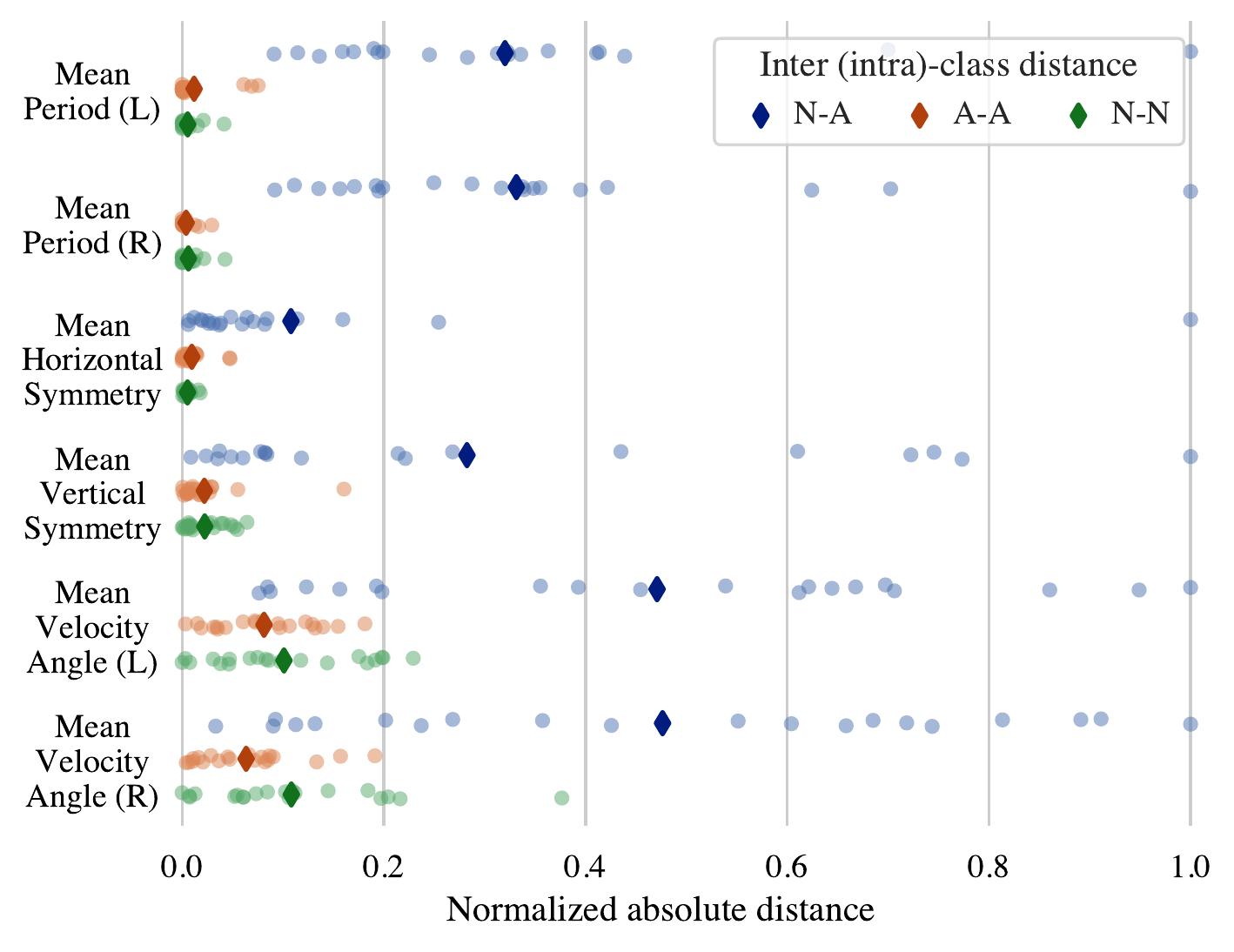}}
    \caption{The inter-class and intra-class distances between some features of normal \textit{(N)} and abnormal \textit{(A)} FTF recordings.
    \ding{169} denotes the mean value.} 
    \label{fig:diff_dist_als}
\end{figure}

Fig.~\ref{fig:diff_dist_als} illustrates the distribution of A-A, N-N, and N-A distances across $20$ different subjects for a subset of features of FTF test. While the intra-class values are concentrated near zero, the inter-class distances are spread out over a wider range. In addition, the mean A-A and N-N distances are strictly lower than the N-A distances. The higher concentration of A-A and N-N distances around zero shows that our feature set is robust to some minor changes in the viewpoint and is not affected by the recording device. Furthermore, it can be seen as a proof-of-concept, demonstrating the ability to compare the subject's performance across different time points. This validates the potential of neurological disease progression the effects of treatment tracking using our DNE system.

\vspace{-10pt}
\subsection{Abnormality Detection}
\vspace{-2pt}
\begin{figure}
\centering
\pgfplotsset{every x tick label/.append style={font=\tiny, yshift=-0.1ex}}
\pgfplotsset{every y tick label/.append style={font=\tiny, xshift=-0.1ex}}

\includegraphics[]{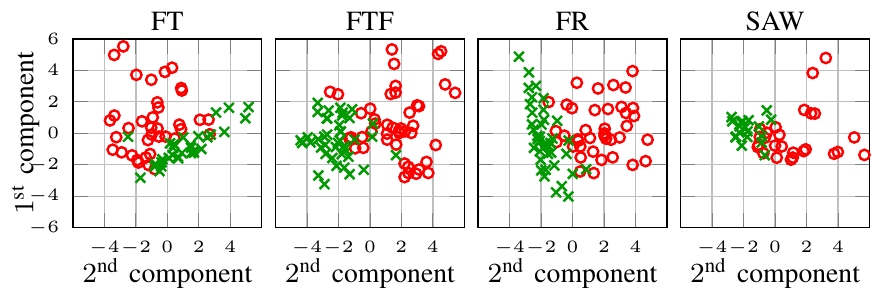}
\caption{PCA analysis of FT, FTF, FR, and SAW tests. \textcolor{forestgreen}{Green} crosses and \textcolor{red}{red} circles stand for normal and abnormal recordings. All subplots share the same axis.}
\label{fig:pca-analysis}
\end{figure}

\begin{figure}
\includegraphics[]{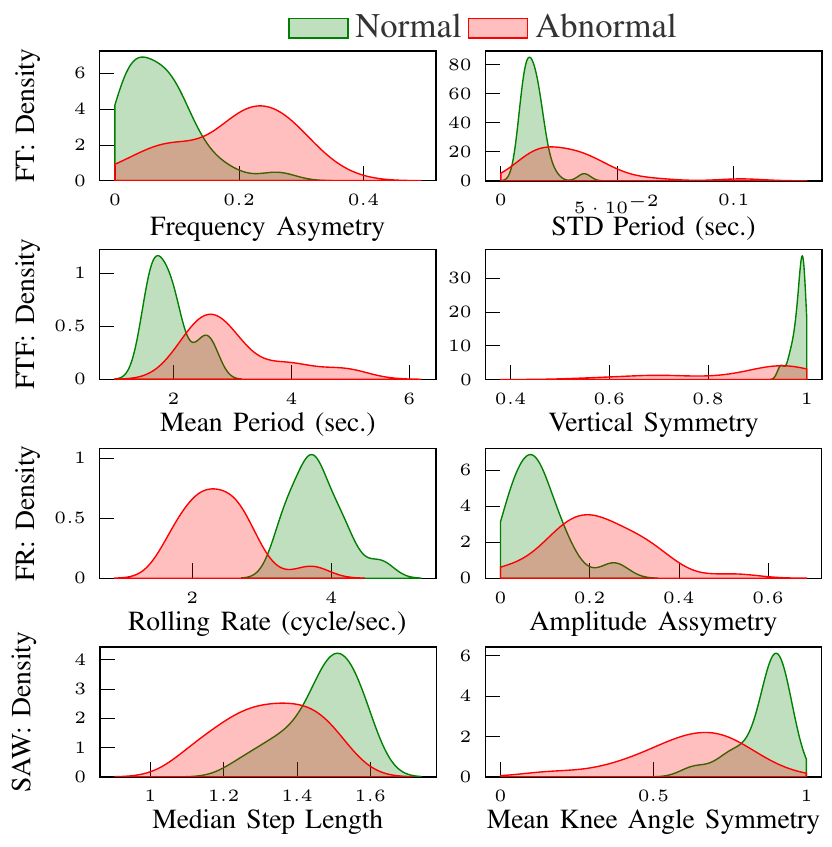}
\caption{Distribution of normal/abnormal features for FT, FTF, FR, SAW tests plotted in first, second, third and last rows. We used kernel density estimation to fit distributions to the data.}
\label{fig:features_density}
\end{figure}

\noindent \textbf{Principal component analysis (PCA):} The feature set describing normal and abnormal recordings constitutes a high-dimensional vector. For a visual comparison of normal and abnormal recordings in terms of their derived features, we perform dimensionality reduction through PCA. For this purpose, for each test, we concatenate the set of features listed in Table~\ref{tab:dne-features} and normalize them before passing to PCA. Fig.~\ref{fig:pca-analysis} showcases the results for different tests. It is observed that the normal and abnormal recordings are separated in dimension reduced feature space. 
This implies that our defined features are descriptive and well differentiate normal from abnormal.

\noindent \textbf{Abnormal Class Distribution:} In Fig.~\ref{fig:features_density} we compare the distribution of normal versus abnormal features for FT, FTF, FR, and SAW tests. These plots clearly indicate the difference in distribution between two classes. Normal features are more concentrated in a specific range, however the abnormal features are often less regular and have a higher STD.    

\noindent \textbf{Abnormality Detection:} We assess the normal and abnormal classification performance using our proposed set of features. Therefore, we utilize several machine learning (ML) models that are grouped into: 1) tree-based methods such as Random Forest (RF), Gradient-Boosting Machine (GBM) \cite{friedman2001greedy}, XGBoost \cite{Chen:2016:XST:2939672.2939785} and 2) parametric models trained using gradient-descent updates, including Logistic Regression (LR),  Support Vector Machine with radial basis function (RBF) kernel (RSVM) and Multi-layer Perceptron (MLP) with rectified linear unit (ReLU) activation. 

We also benchmarked our ML classification performance against two deep learning (DL) baselines. Both DL models predict normal versus abnormal based on major keypoint pose sequence, unlike the ML based models which perform classification on the extracted spatio-temporal/kinematic features. In the first DL baseline, we adopt a long-short term memory (LSTM)~\cite{hochreiter1997long} based sequential model while in the second DL approach, similar to~\cite{kidzinski_automatic_2019}, we use convolutional neural networks (CNN). Details of ML and DL based classification models, data processing and hyper parameters are provided in the supplementary section II and Table S1. We evaluate different models via metrics such as accuracy, average precision, F1 score, and area under the ROC curve (AUC).

We have two splitting schemes to separate the train from test sets in our experiments. In \textit{video-based} splitting, videos from all subjects are divided independently based on a $80 \%$/$20 \%$ splitting ratio for train/test sets. In addition, to evaluate the performance of the models on unseen patients, the \textit{subject-based} division scheme splits a portion of the subjects into the train set while keeping the rest in the test set. Thus, videos belonging to the subjects in the train set are not used in the test set and vice versa. In subject-based splitting, we have $16$/$4$ subjects in train/test sets.

We performed $5$-fold cross validation and summarized the average classification performance of all ML and DL models
in Table \ref{tab:classification}. While all models perform well for various tests, among ML models RSVM and GBM/XGBoost tend to perform better on most metrics. However, the gap between the performance of all ML models is not significant across different tests. This suggests that the extracted set of features well-distinguish normal from abnormal samples.

Furthermore, comparing ML and DL models, we notice that: 1) While DL models perform well on FT, FR and FTF tasks (especially for video-based splitting), they are lagging behind ML models for SAW. We attribute this to the fact that SAW involves more complex motion patterns. Therefore, DL models require larger datasets to be able to learn the classification task from the pose data. 2) DL features extracted from the pose data lack clinical interpretability. 3) For subject-based splitting, ML models operating on the spatio-temporal/kinematic features outperform DL models on most metrics. This indicates better generalization capability of our features on unseen subjects compared to DL models operating on pose data.

\newcommand\NROWS{8}
\begin{table*}
\caption{Classification performance of several machine learning models, including Random Forest (RF), Gradient-BoostingMachine (GBM), XGBoost, LogisticRegression (LR), Support Vector Machine with RBF kernel (RSVM), and Multi-layer Perceptron (MLP) and LSTM and CNN based deep learning models for FT, FTF, FR and SAW tests. The best and second best results are in \textbf{bold} and \underline{underline}, respectively.
}
\vspace{-5pt}
\label{tab:classification}
\centering
\resizebox{\linewidth}{!}{
\pgfplotstabletypeset[
    every head row/.style={
        before row={%
            \toprule
            &&\multicolumn{7}{c}{\textbf{Subject Based}} & \multicolumn{7}{c}{\textbf{Video Based}}\\
            \cmidrule(lr){3-9}
            \cmidrule(lr){10-16}
        },
        after row = \midrule
    },
    columns/0/.style={
        column name={\textbf{Test}},
        assign cell content/.code={
        \pgfmathparse{int(mod(\pgfplotstablerow,\NROWS))}
        \ifnum\pgfmathresult=0
            \pgfkeyssetvalue{/pgfplots/table/@cell content}
            {\multirow{\NROWS}{*}{##1}}%
        \else
            \pgfkeyssetvalue{/pgfplots/table/@cell content}{}%
        \fi
        },
    },
    columns/1/.style ={column name={\textbf{Model}}},
    columns/2/.style={column name={\textbf{Acc}}},
    columns/3/.style={column name={\textbf{Precision}}},
    columns/4/.style={column name={\textbf{Recall}}},
    columns/5/.style={column name={\textbf{Specificity}}},
    columns/6/.style={column name={\textbf{F1 Score}}},
    columns/7/.style={column name={\textbf{AUC}}},
    columns/8/.style={column name={\textbf{AP}}, 
    column type/.add={}{|}},
    columns/9/.style={column name={\textbf{Acc}}},
    columns/10/.style={column name={\textbf{Precision}}},
    columns/11/.style={column name={\textbf{Recall}}},
    columns/12/.style={column name={\textbf{Specificity}}},
    columns/13/.style={column name={\textbf{F1 Score}}},
    columns/14/.style={column name={\textbf{AUC}}},
    columns/15/.style={column name={\textbf{AP}}},
    every row no 6/.style={before row=\cmidrule(lr){2-16}},
    every row no 14/.style={before row=\cmidrule(lr){2-16}},
    every row no 22/.style={before row=\cmidrule(lr){2-16}},
    every row no 30/.style={before row=\cmidrule(lr){2-16}},
    every nth row={\NROWS}{before row=\midrule},
    header=false,
    col sep=comma,
    row sep=\\,
    every last row/.style={
        after row=\bottomrule},
    string type,
]{
FT,RF,0.8554,0.8947,0.8500,0.8750,0.8500,0.8625,0.8339,0.8773,0.9278,0.8492,0.9236,0.8672,0.8864,0.8643\\
FT,GBM,\textbf{0.8804},0.9156,\textbf{0.8750},0.9000,\textbf{0.8742},\textbf{0.8875},\underline{0.8602},0.8866,0.9464,0.8470,\textbf{0.9418},0.8839,0.8944,\underline{0.8864}\\
FT,XGBOOST,0.8304,0.8778,0.8250,0.8500,0.8263,0.8375,0.8049,0.8655,0.9206,0.8292,0.9218,0.8481,0.8755,0.8530\\
FT,LR,0.8679,\textbf{0.9714},0.7750,\textbf{0.9750},0.8514,\underline{0.8750},\textbf{0.8732},0.8773,\underline{0.9492},0.8292,\textbf{0.9418},0.8613,0.8855,0.8775\\
FT,RSVM,0.8679,0.8950,\textbf{0.8750},0.8750,\underline{0.8639},\underline{0.8750},0.8428,\underline{0.8916},0.9014,\textbf{0.8914},0.8951,\underline{0.8867},0.8932,0.8631\\
FT,MLP,0.8679,\underline{0.9350},0.8250,\underline{0.9250},0.8575,\underline{0.8750},0.8578,0.8563,0.9300,0.8029,0.9236,0.8199,0.8632,0.8387\\
FT,LSTM,0.8089,0.8273,0.8250,0.8000,0.8146,0.8125,0.7705,\textbf{0.9008},\underline{0.9492},\underline{0.8796},\textbf{0.9418},\textbf{0.9029},\textbf{0.9107},\textbf{0.9044}\\
FT,CNN,0.8304,0.8273,\textbf{0.8750},0.7833,0.8474,0.8292,0.8024,\underline{0.8916},\textbf{0.9514},0.8514,\textbf{0.9418},0.8730,\underline{0.8966},0.8852\\
FTF,RF,0.8625,0.9232,0.8250,0.9000,0.8510,0.8625,0.8357,0.9623,0.9550,\underline{0.9818},0.9400,0.9666,0.9609,0.9473\\
FTF,GBM,\underline{0.9125},0.9378,\underline{0.9000},0.9250,\underline{0.8993},\underline{0.9125},\underline{0.8878},\textbf{0.9895},\underline{0.9800},\textbf{1.0000},\underline{0.9800},\textbf{0.9895},\textbf{0.9900},\textbf{0.9800}\\
FTF,XGBOOST,\textbf{0.9250},0.9278,\textbf{0.9250},0.9250,\textbf{0.9249},\textbf{0.9250},\textbf{0.9028},0.9684,\underline{0.9800},0.9636,\underline{0.9800},0.9704,0.9718,0.9647\\
FTF,LR,0.8375,0.9378,0.7500,0.9250,0.8004,0.8375,0.8128,0.8930,0.9314,0.8805,0.9200,0.8988,0.9003,0.8853\\
FTF,RSVM,0.8875,0.9378,0.8500,0.9250,0.8708,0.8875,0.8628,0.9579,0.9600,0.9636,0.9600,0.9599,0.9618,0.9447\\
FTF,MLP,0.8625,0.8788,0.8750,0.8500,0.8619,0.8625,0.8218,\underline{0.9789},0.9778,0.9778,\underline{0.9800},0.9778,0.9789,0.9686\\
FTF,LSTM,0.8875,\textbf{1.0000},0.7750,\textbf{1.0000},0.8338,0.8875,0.8875,\underline{0.9789},\underline{0.9800},\underline{0.9818},\underline{0.9800},\underline{0.9799},\underline{0.9809},0.9723\\
FTF,CNN,0.8875,\underline{0.9492},0.8250,\underline{0.9500},0.8735,0.8875,0.8688,0.9684,\textbf{1.0000},0.9455,\textbf{1.0000},0.9705,0.9727,\underline{0.9770}\\
FR,RF,0.8250,0.9100,0.7500,0.9000,0.8040,0.8250,0.7975,0.8737,0.8656,0.8583,0.8873,0.8551,0.8728,0.8093\\
FR,GBM,0.8500,0.8878,0.8250,0.8750,0.8360,0.8500,0.8128,\underline{0.9033},0.9124,\underline{0.8806},0.8936,\underline{0.8914},0.8871,\underline{0.8543}\\
FR,XGBOOST,0.8500,0.9100,0.8000,0.9000,0.8325,0.8500,0.8225,0.8947,0.9064,0.8583,0.9255,0.8742,\underline{0.8919},0.8406\\
FR,LR,0.8625,\textbf{0.9500},0.7750,\textbf{0.9500},0.8414,0.8625,0.8500,0.8947,\underline{0.9492},0.8083,\underline{0.9618},0.8635,0.8851,0.8513\\
FR,RSVM,\textbf{0.9125},\underline{0.9278},\textbf{0.9000},\underline{0.9250},\textbf{0.9097},\textbf{0.9125},\textbf{0.8903},0.8717,0.8850,0.8417,0.9055,0.8567,0.8736,0.8221\\
FR,MLP,0.7875,0.8955,0.7000,0.8750,0.7413,0.7875,0.7580,0.8132,0.8337,0.8000,0.7891,0.8084,0.7945,0.7605\\
FR,LSTM,\underline{0.8875},0.9100,\underline{0.8750},0.9000,\underline{0.8859},\underline{0.8875},\underline{0.8600},0.8507,0.8929,0.7667,0.9255,0.8227,0.8461,0.7959\\
FR,CNN,0.8000,0.8700,0.7250,0.8750,0.7761,0.8000,0.7650,\textbf{0.9539},\textbf{1.0000},\textbf{0.9222},\textbf{1.0000},\textbf{0.9568},\textbf{0.9611},\textbf{0.9683}\\
SAW,RF,0.7877,0.8167,0.7917,0.7946,0.7804,0.7932,0.7542,0.8000,0.8679,0.8429,0.7467,0.8385,0.7948,\underline{0.8270}\\
SAW,GBM,0.8189,\textbf{0.9000},0.7917,\textbf{0.8571},0.8042,0.8244,0.7958,\underline{0.8200},0.8406,\underline{0.9000},0.6967,0.8561,\underline{0.7983},0.8139\\
SAW,XGBOOST,0.8261,0.8250,\underline{0.8542},0.8036,0.8240,0.8289,0.7677,\underline{0.8200},0.8317,\textbf{0.9333},0.6300,\underline{0.8670},0.7817,0.8106\\
SAW,LR,0.8189,0.8375,\underline{0.8542},0.7946,0.8250,0.8244,0.7802,0.7800,\underline{0.8762},0.7810,\textbf{0.7867},0.8097,0.7838,0.8257\\
SAW,RSVM,\textbf{0.8606},0.8500,\textbf{0.9375},0.7946,\textbf{0.8740},\textbf{0.8661},\underline{0.8187},\textbf{0.8400},\textbf{0.8929},0.8714,\textbf{0.7867},\textbf{0.8685},\textbf{0.8290},\textbf{0.8514}\\
SAW,MLP,0.8189,0.8500,\underline{0.8542},0.7946,0.8240,0.8244,0.7771,0.7800,0.8179,0.8714,0.6467,0.8277,0.7590,0.7860\\
SAW,LSTM,0.7372,0.8333,0.6250,0.8393,0.6778,0.7321,0.6950,0.7800,0.7833,0.8648,0.6700,0.8139,0.7674,0.7541\\
SAW,CNN,0.7877,0.8542,0.7292,0.8393,0.7643,0.7842,0.7452,0.7800,0.8267,0.8076,0.7500,0.8063,0.7788,0.7962\\
}
}
\vspace{-4mm}
\end{table*}
\vspace{-10pt}
\subsection{Feature Importance Analysis}
\vspace{-2pt}
One benefit of tree-based models is in the tractable decision-making process. Therefore, we investigate the importance of each feature, contributing to the decision process by analyzing our RF models. This analysis gives us the weight of all features, sorted in descending order in Supplementary Fig. S4. %

We notice that symmetry between specific R/L features for FT, FTF, and SAW tests is considered the most important, i.e., with the largest weight. For the SAW test, the most important feature is the similarity between the knee angle time series across different cycles ({${S}_{\text{saw}}^{\text{knee angle}}$}) while
for FT (Supplementary Fig. S4-(a)) and FTF (Supplementary Fig. S4-(c)), the features with the largest weights are frequency asymmetry
and horizontal ({${S}_{\text{ftf}}^{\text{finger-x}}$}) symmetry, respectively. Although this can be attributed to the nature of the simulated impairments in our dataset, it is consistent with the clinical practice, where the left and right asymmetry is a common biomarker~\cite{Babrak2019, Nagasaki2004AsymmetricVA, sawyer_asymmetry_1993} of different neurological disorders.

Furthermore, temporal and spatial features that characterize the periodic behavior of the movement are important metrics that the decision tree classification models rely on. Examples of these features are amplitude and period for FT, FTF, and FR tests, step length, width, and step time for SAW.
We also notice that for a subset of features, having large variations (i.e. STD) across different cycles is another indicator of abnormal performance in our dataset. This is captured in the large weight associated with STD values of some features for various tests. This result also affirms our observations in Fig.~\ref{fig:features_density}.

\vspace{-10pt}
\section{Discussion \& Challenges}
\vspace{-5pt}
In this section, we discuss various aspects of DNE including feature design, robustness, clinical relevance and application as well as the current challenges and our proposed solutions.

\vspace{-10pt}
\subsection{Discussion}
\vspace{-2pt}
\textbf{Feature Design:} The main goal of our DNE system is to provide an objective tool for quantifying and documentation of  recordings of neurological tests. Thus, it is critical to design a set of clinical interpretable features that explain the performance of a subject on various motor tasks. In addition, having powerful digital biomarkers reduces the workload of normal versus abnormal classification models and improves their generalization, especially when large training datasets are not available. Furthermore, unlike black-box DL models, the explainability of our diverse set of features allows clinicians to better understand and track patients' status over time.   

\textbf{Robustness:} DNE is resilient to changes in slight deviations from the camera view, distance to the camera, subject clothing, and mild pixel intensity changes due to intermediate data standardization and robust pose estimation steps (section \ref{ssec:pose-estimation}-\ref{sec:prep}). This is experimentally shown by the low intra-class feature distances in Fig. \ref{fig:diff_dist_als}. Data normalization and filtering in the pre-processing step also helps in eliminating noise and propagated errors from the pose estimation module.

In FT, FR and SAW tests, the abnormality in the motion is imposed by wearing equipment which are visible in the recordings. The pose estimation models we have used (OP and VP3D) are robust to the appearance of the equipment and can accurately predict the joint locations regardless of the presence of the equipment. The features
incorporated in the classification tasks are derived from the pose data. Therefore, the quantified features and the classification performance is not affected by the visual cues from the equipment.

\textbf{Clinical Relevance:} In our dataset, the abnormalities in the movements of the subjects were simulated. The simulated impairment in the FR test is the closest to what is observed in clinics for patients with neurological disorders. In the simulated impairment for FR, the arm with no moulage satellites around the weighted wrist, causing a decrease in the orbit frequency (Fig.~\ref{fig:fr_example}). This is coherent with the clinical observations of patients with neurological impairments. 

In the FTF test, the simulated abnormality would be more realistic, if the tremor or inaccuracy of movement increased as the finger got closer to its target (i.e. when the two fingers approach). In our current dataset, the subjects often simulated the tremor throughout their movements which is only seen in severe cases. In addition, for the FT test, often the abnormality is a combination of decreased amplitude and rate (Fig.~\ref{fig:ft_features}) and in Parkinson’s decrements of both. In our DNE dataset, some subjects simulated more of one or the other.

In SAW, the abnormality in real patients appear as a combination of slow time-to-stand, decreased step length, increased step-time, and asymmetry of gait features. In our dataset, the abnormality was imposed by wearing a knee brace. Alongside asymmetry between the R/L knee angles, we observed decreased step-length for the subjects wearing the knee brace (Fig. ~\ref{fig:features_density}-SAW). These are in-line with clinical observations from real patients.

Overall, features that clinicians observe were disrupted from normal findings to various degrees, although the pattern of disruption of features may have not been exact for a specific condition. We showed that DNE was able to define clinically interpretable features and detect differences between normal and simulated impaired recordings.
As future work, to expand its clinical impact, we will focus our analysis on real patients with various neurological impairment severity levels, 
and with other neurological tests, such as eye movement~\cite{Pretegiani2017}, facial activation~\cite{Jin2020,Rafayet2021}, or phonation \cite{Fabbri2017}.

\textbf{Clinical Application:} The initial clinical application of DNE is measuring and documenting features of various neurological exams. This would allow for improved communication of objective exam quantification and the ability to assess for changes over time. As future work, with clinicians' supervision, we will examine and report the performance of DNE on real patients. A longer term goal is to assist clinicians with classification of recordings and provide a platform for longitudinal monitoring of patients.

\vspace{-10pt}
\subsection{Challenges}
\vspace{-2pt}
\textbf{Depth Ambiguity: }Analyzing human motion from 2D RGB data requires dealing with uncertainties associated with lacking depth information. Furthermore, depth ambiguity becomes a more prominent challenge for the SAW test with frontal view recordings rather than sagittal view. It also avoids defining the spatial features in their absolute units. Currently, to mitigate the issues corresponding to these depth uncertainties, for upper limb tests, the subjects are asked to perform the tests while facing the camera and (roughly) in parallel to its image plane. In our processing steps, we also perform pose normalizations to compensate for scale variations due to variable distance from the camera. To further address this issue, we believe incorporating LiDAR depth maps captured by recent iOS devices, in the pose estimation step can prove helpful.

\textbf{Self-baselining:} Natural motion properties differ across various subjects. For example, one subject can be inherently slower or have less strength in performing some tests. In our dataset, we witnessed while some subjects had a slower inherent speed in their normal performance, they were mistakenly classified as abnormal. This highlights the importance of taking into account the history of a subject and self-baselining. 
In our experiments, we showcased an example of self-comparisons of normal and abnormal performance of the same subject (Fig.~\ref{fig:diff_dist_als}). The purpose of this study was to show the ability of our designed features to discriminate between the varying status of the subject at different test times. This result validates the potential of our DNE pipeline as a personalized medical assessment system that helps clinicians better monitor the disease progression and the effect of medical treatments.

\textbf{Real-time DNE:} Our current DNE system and the extracted kinematic/spatio temporal features rely on tracking the human pose from the video recordings in an offline step using off-the-shelf pose estimation modules. Currently, the pose estimation step is the most computationally expensive step, hindering real time processing and feature extraction. To address this challenge, on-device lighter pose estimation models (with small sacrifice on the accuracy), that focus on extracting major keypoints rather than the whole body pose are necessary.

\vspace{-10pt}
\section{Conclusion}
\vspace{-2pt}
\label{sec:conclusion}

In this paper, we proposed a comprehensive vision-based digital biomarker exam solution named Digitized Neurological Examination (DNE). Using DNE software, users video record their performance on various motor tasks, including finger tapping, finger to finger, forearm roll, and stand-up and walk. We introduced the DNE dataset, a total of $361$ videos consisting of normal and impaired functions of $21$ subjects, performing different tests.
For each recording, 2D/3D pose is estimated and used to quantify kinematic and spatio-temporal features. These features form a set of digital biomarkers that can be 1) accurately obtained from common RGB videos with minimal calibration, 2) used to track the clinical changes across recordings at different time points.  
On our DNE dataset, we analyzed the effectiveness of the defined features in differentiating normal versus impaired simulated videos per and across subjects. Our results demonstrate high classification accuracy and F1 scores using a variety of machine learning models. Future work will extend the setting of this study to a larger set of subjects with a diverse range of abnormalities.

\vspace{-8pt}
\section*{Acknowledgement}
\vspace{-3pt}
This work was supported by the Jump ARCHES endowment through the Health Care Engineering Systems Center, by the National Institute of Health (NIH) R01 AI139401, and by the Vingroup Innovation Foundation VINIF.2021.DA00128.

\vspace{-8pt}
\small{
\bibliographystyle{IEEEtran}
\bibliography{refs.bib}}

\begin{thebibliography}{10}
\providecommand{\url}[1]{#1}
\csname url@samestyle\endcsname
\providecommand{\newblock}{\relax}
\providecommand{\bibinfo}[2]{#2}
\providecommand{\BIBentrySTDinterwordspacing}{\spaceskip=0pt\relax}
\providecommand{\BIBentryALTinterwordstretchfactor}{4}
\providecommand{\BIBentryALTinterwordspacing}{\spaceskip=\fontdimen2\font plus
\BIBentryALTinterwordstretchfactor\fontdimen3\font minus
  \fontdimen4\font\relax}
\providecommand{\BIBforeignlanguage}[2]{{%
\expandafter\ifx\csname l@#1\endcsname\relax
\typeout{** WARNING: IEEEtran.bst: No hyphenation pattern has been}%
\typeout{** loaded for the language `#1'. Using the pattern for}%
\typeout{** the default language instead.}%
\else
\language=\csname l@#1\endcsname
\fi
#2}}
\providecommand{\BIBdecl}{\relax}
\BIBdecl

\bibitem{Feigin_2021}
V.~L. Feigin \emph{et~al.}, ``Burden of neurological disorders across the us
  from 1990-2017: A global burden of disease study.'' \emph{JAMA neurology},
  vol.~78, no.~2, pp. 165 -- 176, 2021.

\bibitem{Dall_2013}
T.~Dall \emph{et~al.}, ``Supply and demand analysis of the current and future
  us neurology workforce.'' \emph{Neurology}, vol.~81, no.~5, pp. 470--478,
  2013.

\bibitem{grossman_rapid_2020}
S.~N. Grossman \emph{et~al.}, ``\BIBforeignlanguage{en}{Rapid implementation of
  virtual neurology in response to the {COVID}-19 pandemic},''
  \emph{\BIBforeignlanguage{en}{Neurology}}, vol.~94, no.~24, pp. 1077--1087,
  Jun. 2020.

\bibitem{al_hussona_virtual_2020}
M.~Al~Hussona \emph{et~al.}, ``The {Virtual} {Neurologic} {Exam}:
  {Instructional} {Videos} and {Guidance} for the {COVID}-19 {Era},'' \emph{The
  Canadian Journal of Neurological Sciences.}, pp. 1--6, May 2020.

\bibitem{capozzo_telemedicine_2020}
R.~Capozzo \emph{et~al.}, ``\BIBforeignlanguage{eng}{Telemedicine is a useful
  tool to deliver care to patients with {Amyotrophic} {Lateral} {Sclerosis}
  during {COVID}-19 pandemic: results from {Southern} {Italy}},''
  \emph{\BIBforeignlanguage{eng}{Amyotrophic Lateral Sclerosis \&
  Frontotemporal Degeneration}}, vol.~21, no. 7-8, pp. 542--548, Nov. 2020.

\bibitem{duncan_video_2020}
C.~Duncan \emph{et~al.}, ``\BIBforeignlanguage{en}{Video consultations in
  ordinary and extraordinary times},'' \emph{\BIBforeignlanguage{en}{Practical
  Neurology}}, vol.~20, no.~5, pp. 396--403, Oct. 2020.

\bibitem{patterson_neurological_2021}
V.~Patterson, ``\BIBforeignlanguage{en}{Neurological telemedicine in the
  {COVID}-19 era},'' \emph{\BIBforeignlanguage{en}{Nature Reviews Neurology}},
  vol.~17, no.~2, pp. 73--74, Feb. 2021.

\bibitem{Mutgi_2015}
S.~A. Mutgi \emph{et~al.}, ``Emerging subspecialties in neurology: Telestroke
  and teleneurology.'' \emph{Neurology}, vol.~84, no.~22, pp. 191 -- 193, 2015.

\bibitem{Dorsey_2018}
E.~R. Dorsey \emph{et~al.}, ``Teleneurology and mobile technologies: the future
  of neurological care.'' \emph{Nature Reviews Neurology}, vol.~14, no.~5, pp.
  285 -- 297, 2018.

\bibitem{Adams2021}
J.~L. Adams \emph{et~al.}, ``Digital technology in movement disorders: Updates,
  applications, and challenges,'' \emph{Current neurology and neuroscience
  reports}, vol.~21, no.~4, pp. 16--16, Mar. 2021.

\bibitem{Williams2020}
S.~Williams \emph{et~al.}, ``The discerning eye of computer vision: Can it
  measure {Parkinson}'s finger tap bradykinesia?'' \emph{Journal of the
  Neurological Sciences}, vol. 416, p. 117003, 2020.

\bibitem{pmlr-v68-jaroensri17a}
R.~Jaroensri \emph{et~al.}, ``A video-based method for automatically rating
  ataxia,'' in \emph{Proceedings of the 2nd Machine Learning for Healthcare
  Conference}, vol.~68, Aug. 2017, pp. 204--216.

\bibitem{Xue2018}
D.~Xue \emph{et~al.}, ``Vision-based gait analysis for senior care,''
  \emph{CoRR}, vol. abs/1812.00169, 2018.

\bibitem{tedim_cruz_novel_2014}
V.~Tedim~Cruz \emph{et~al.}, ``\BIBforeignlanguage{en}{A novel system for
  automatic classification of upper limb motor function after stroke: {An}
  exploratory study},'' \emph{\BIBforeignlanguage{en}{Medical Engineering \&
  Physics}}, vol.~36, no.~12, pp. 1704--1710, Dec. 2014.

\bibitem{einsler_sarahome_2021}
M.~Grobe-Einsler \emph{et~al.}, ``Development of sarahome, a new video-based
  tool for the assessment of ataxia at home,'' \emph{Movement Disorders},
  vol.~36, no.~5, pp. 1242--1246, 2021.

\bibitem{Wei2021}
Y.~Wei \emph{et~al.}, ``Interactive video acquisition and learning system for
  motor assessment of parkinson's disease,'' in \emph{Proceedings of the
  Thirtieth International Joint Conference on Artificial Intelligence,
  {IJCAI-21}}, Z.-H. Zhou, Ed.\hskip 1em plus 0.5em minus 0.4em\relax
  International Joint Conferences on Artificial Intelligence Organization, 8
  2021, pp. 5024--5027, demo Track.

\bibitem{Cao2019_openpose}
Z.~{Cao} \emph{et~al.}, ``Openpose: Realtime multi-person {2D} pose estimation
  using part affinity fields,'' \emph{IEEE Transactions on Pattern Analysis and
  Machine Intelligence}, 2019.

\bibitem{pavllo2019_videopose3d}
D.~Pavllo \emph{et~al.}, ``3{D} human pose estimation in video with temporal
  convolutions and semi-supervised training,'' in \emph{Conference on Computer
  Vision and Pattern Recognition (CVPR)}, 2019.

\bibitem{john_biomechanics}
\emph{Kinematics}.\hskip 1em plus 0.5em minus 0.4em\relax John Wiley \& Sons,
  Ltd, 2009, ch.~3, pp. 45--81.

\bibitem{jovicic2014}
M.~Djurić-Jovičić \emph{et~al.}, ``Implementation of continuous wavelet
  transformation in repetitive finger tapping analysis for patients with pd,''
  in \emph{2014 22nd Telecommunications Forum Telfor (TELFOR)}, 2014, pp.
  541--544.

\bibitem{Yokoe2009}
M.~Yokoe \emph{et~al.}, ``Opening velocity, a novel parameter, for finger
  tapping test in patients with parkinson's disease,'' \emph{Parkinsonism and
  Related Disorders}, vol.~15, no.~6, pp. 440--444, 2009.

\bibitem{Jia2014}
X.~Jia \emph{et~al.}, ``Objective quantification of upper extremity motor
  functions in unified {Parkinson's Disease Rating Scale} test,'' in \emph{2014
  36th Annual International Conference of the IEEE Engineering in Medicine and
  Biology Society}, 2014, pp. 5345--5348.

\bibitem{Lee2016}
C.~Lee \emph{et~al.}, ``A validation study of a smartphone-based finger tapping
  application for quantitative assessment of bradykinesia in parkinson's
  disease,'' \emph{PloS one}, vol.~11, p. e0158852, Jul. 2016.

\bibitem{Memedi2015}
M.~Memedi \emph{et~al.}, ``Validity and {Responsiveness} of {A}t-home {T}ouch
  {S}creen {A}ssessments in {Advanced Parkinson's Disease},'' \emph{IEEE
  Journal of Biomedical and Health Informatics}, vol.~19, no.~6, pp.
  1829--1834, 2015.

\bibitem{Aghanavesi2017}
S.~Aghanavesi \emph{et~al.}, ``A smartphone-based system to quantify dexterity
  in {Parkinson}'s disease patients,'' \emph{Informatics in Medicine Unlocked},
  vol.~9, pp. 11--17, 2017.

\bibitem{Zhang2018}
A.~Zhan \emph{et~al.}, ``Using smartphones and machine learning to quantify
  {P}arkinson disease severity: The mobile {P}arkinson disease score,''
  \emph{JAMA neurology}, vol.~75, no.~7, pp. 876--880, Jul. 2018.

\bibitem{Jobbagy2005}
Ákos Jobbágy \emph{et~al.}, ``Analysis of finger-tapping movement,''
  \emph{Journal of Neuroscience Methods}, vol. 141, no.~1, pp. 29--39, 2005.

\bibitem{Khan2014}
T.~Khan \emph{et~al.}, ``A computer vision framework for finger-tapping
  evaluation in parkinson's disease,'' \emph{Artificial Intelligence in
  Medicine}, vol.~60, no.~1, pp. 27--40, 2014.

\bibitem{Liu2019}
Y.~Liu \emph{et~al.}, ``Vision-based method for automatic quantification of
  {Parkinsonian Bradykinesia},'' \emph{IEEE Transactions on Neural Systems and
  Rehabilitation Engineering}, vol.~27, no.~10, pp. 1952--1961, 2019.

\bibitem{Li2021}
H.~Li \emph{et~al.}, ``Automated assessment of parkinsonian finger-tapping
  tests through a vision-based fine-grained classification model,''
  \emph{Neurocomputing}, vol. 441, pp. 260--271, 2021.

\bibitem{rodrigues_chronic_stoke_2017}
M.~R.~M. Rodrigues \emph{et~al.}, ``Does the {Finger}-to-{Nose} {Test} measure
  upper limb coordination in chronic stroke?'' \emph{Journal of
  NeuroEngineering and Rehabilitation}, vol.~14, Jan. 2017.

\bibitem{oubre_decomposition_2021}
B.~Oubre \emph{et~al.}, ``\BIBforeignlanguage{eng}{Decomposition of {Reaching}
  {Movements} {Enables} {Detection} and {Measurement} of {Ataxia}},''
  \emph{\BIBforeignlanguage{eng}{Cerebellum (London, England)}}, Mar. 2021.

\bibitem{gajos_computer_mouse_2020}
K.~Z. Gajos \emph{et~al.}, ``Computer mouse use captures {Ataxia} and
  {Parkinsonism}, enabling accurate measurement and detection,'' \emph{Movement
  Disorders}, vol.~35, no.~2, pp. 354--358, 2020.

\bibitem{Simonsen2017}
D.~Simonsen \emph{et~al.}, ``\BIBforeignlanguage{English}{Design and test of a
  {M}icrosoft {K}inect-based system for delivering adaptive visual feedback to
  stroke patients during training of upper limb movement},''
  \emph{\BIBforeignlanguage{English}{Medical \& Biological Engineering \&
  Computing}}, vol.~55, no.~11, p. 1927–1935, 2017.

\bibitem{Hoffmann_2007}
T.~Hoffmann \emph{et~al.}, ``Remote measurement via the internet of upper limb
  range of motion in people who have had a stroke.'' \emph{Journal of
  Telemedicine and Telecare}, vol.~13, no.~8, pp. 401--405, 2007.

\bibitem{allin_robust_2010}
S.~Allin \emph{et~al.}, ``\BIBforeignlanguage{en}{Robust {Tracking} of the
  {Upper} {Limb} for {Functional} {Stroke} {Assessment}},''
  \emph{\BIBforeignlanguage{en}{IEEE Transactions on Neural Systems and
  Rehabilitation Engineering}}, p.~9, 2010.

\bibitem{kour_computer-vision_2019}
N.~Kour \emph{et~al.}, ``Computer-{Vision} {Based} {Diagnosis} of
  {Parkinson}’s {Disease} via {Gait}: {A} {Survey},'' \emph{IEEE Access},
  vol.~7, pp. 156\,620--156\,645, 2019.

\bibitem{ortells_vision-based_2018}
M.~R. Ortells~J, Herrero-Ezquerro~MT,
  ``\BIBforeignlanguage{English}{Vision-based gait impairment analysis for
  aided diagnosis},'' \emph{\BIBforeignlanguage{English}{Medical and Biological
  Engineering and Computing}}, vol.~56, no.~9, pp. 1553--1564, 2018.

\bibitem{nieto-hidalgo_gait_2018}
M.~Nieto-Hidalgo \emph{et~al.}, ``\BIBforeignlanguage{en}{Gait analysis using
  computer vision based on cloud platform and mobile device},'' Jan. 2018.

\bibitem{zhu_computer_2016}
W.~Zhu \emph{et~al.}, ``\BIBforeignlanguage{en}{A {Computer} {Vision}-{Based}
  {System} for {Stride} {Length} {Estimation} using a {Mobile} {Phone}
  {Camera}},'' in \emph{\BIBforeignlanguage{en}{Proceedings of the 18th
  {International} {ACM} {SIGACCESS} {Conference} on {Computers} and
  {Accessibility}}}.\hskip 1em plus 0.5em minus 0.4em\relax Reno Nevada USA:
  ACM, Oct. 2016, pp. 121--130.

\bibitem{NIETOHIDALGO2016}
M.~Nieto-Hidalgo \emph{et~al.}, ``A vision based proposal for classification of
  normal and abnormal gait using rgb camera,'' \emph{Journal of Biomedical
  Informatics}, vol.~63, pp. 82--89, 2016.

\bibitem{Toshev2013}
A.~Toshev \emph{et~al.}, ``Deeppose: Human pose estimation via deep neural
  networks,'' \emph{Proceedings of the IEEE Computer Society Conference on
  Computer Vision and Pattern Recognition}, Dec. 2013.

\bibitem{Sun_2019_CVPR}
K.~Sun \emph{et~al.}, ``Deep high-resolution representation learning for human
  pose estimation,'' in \emph{Proceedings of the IEEE/CVF Conference on
  Computer Vision and Pattern Recognition (CVPR)}, Jun. 2019.

\bibitem{Martinez_2017_ICCV}
J.~Martinez \emph{et~al.}, ``A simple yet effective baseline for 3{D} human
  pose estimation,'' in \emph{Proceedings of the IEEE International Conference
  on Computer Vision (ICCV)}, Oct. 2017.

\bibitem{Rong2020}
Y.~Rong \emph{et~al.}, ``Frankmocap: A monocular 3d whole-body pose estimation
  system via regression and integration,'' in \emph{IEEE International
  Conference on Computer Vision Workshops}, 2021.

\bibitem{Zimmermann2018}
C.~Zimmermann \emph{et~al.}, ``3{D} human pose estimation in {RGBD} images for
  robotic task learning,'' in \emph{2018 IEEE International Conference on
  Robotics and Automation (ICRA)}, 2018, pp. 1986--1992.

\bibitem{Clark2018}
R.~Clark \emph{et~al.}, ``Three-dimensional cameras and skeleton pose tracking
  for physical function assessment: A review of uses, validity, current
  developments and kinect alternatives,'' \emph{Gait \& Posture}, vol.~68, Nov.
  2018.

\bibitem{Springer2016}
S.~Springer \emph{et~al.}, ``Validity of the kinect for gait assessment: A
  focused review,'' \emph{Sensors}, vol.~16, no.~2, 2016.

\bibitem{Dolatabadi2016}
E.~Dolatabadi \emph{et~al.}, ``Automated classification of pathological gait
  after stroke using ubiquitous sensing technology,'' in \emph{2016 38th Annual
  International Conference of the IEEE Engineering in Medicine and Biology
  Society (EMBC)}, 2016, pp. 6150--6153.

\bibitem{Joao2020}
J.~Andre \emph{et~al.}, ``Markerless gait analysis vision system for real-time
  gait monitoring.'' \emph{2020 IEEE International Conference on Autonomous
  Robot Systems and Competitions (ICARSC)}, pp. 269 -- 274, 2020.

\bibitem{Li2018NS}
T.~Li \emph{et~al.}, ``Automatic timed up-and-go sub-task segmentation for
  parkinson’s disease patients using video-based activity classification,''
  \emph{IEEE Transactions on Neural Systems and Rehabilitation Engineering},
  vol.~26, no.~11, pp. 2189--2199, 2018.

\bibitem{sato_quantifying_2019}
K.~Sato \emph{et~al.}, ``\BIBforeignlanguage{en}{Quantifying normal and
  {parkinsonian} gait features from home movies: {Practical} application of a
  deep learning–based {2D} pose estimator},''
  \emph{\BIBforeignlanguage{en}{PLOS ONE}}, vol.~14, no.~11, p. e0223549, Nov.
  2019.

\bibitem{Hu2020}
K.~Hu \emph{et~al.}, ``Vision-based freezing of gait detection with anatomic
  directed graph representation,'' \emph{IEEE Journal of Biomedical and Health
  Informatics}, vol.~24, no.~4, pp. 1215--1225, 2020.

\bibitem{kidzinski_deep_2020}
L.~Kidzi{\'n}ski \emph{et~al.}, ``\BIBforeignlanguage{en}{Deep neural networks
  enable quantitative movement analysis using single-camera videos},''
  \emph{\BIBforeignlanguage{en}{Nature Communications}}, vol.~11, no.~1, p.
  4054, Dec. 2020.

\bibitem{kimore}
M.~Capecci \emph{et~al.}, ``The {KIMORE} dataset: Kinematic assessment of
  movement and clinical scores for remote monitoring of physical
  rehabilitation,'' \emph{IEEE Transactions on Neural Systems and
  Rehabilitation Engineering}, vol.~27, no.~7, pp. 1436--1448, 2019.

\bibitem{Simon2017}
T.~Simon \emph{et~al.}, ``Hand keypoint detection in single images using
  multiview bootstrapping,'' in \emph{Proceedings of the IEEE Conference on
  Computer Vision and Pattern Recognition (CVPR)}, Jul. 2017.

\bibitem{wu2019detectron2}
Y.~Wu \emph{et~al.}, ``Detectron2,''
  \url{https://github.com/facebookresearch/detectron2}, 2019.

\bibitem{Savitzky1964}
A.~Savitzky \emph{et~al.}, ``Smoothing and differentiation of data by
  simplified least squares procedures.'' \emph{Analytical Chemistry}, vol.~36,
  no.~8, pp. 1627--1639, 1964.

\bibitem{krishna_quantitative_2019}
R.~Krishna \emph{et~al.}, ``Quantitative assessment of cerebellar ataxia,
  through automated limb functional tests,'' \emph{Journal of NeuroEngineering
  and Rehabilitation}, vol.~16, no.~1, p.~31, Feb. 2019.

\bibitem{Jawad2012}
A.~Al-Jawad \emph{et~al.}, ``Using multi-dimensional dynamic time warping for
  tug test instrumentation with inertial sensors,'' in \emph{2012 IEEE
  International Conference on Multisensor Fusion and Integration for
  Intelligent Systems (MFI)}, 2012, pp. 212--218.

\bibitem{Herran2014}
A.~Muro-de-la Herran \emph{et~al.}, ``Gait analysis methods: An overview of
  wearable and non-wearable systems, highlighting clinical applications,''
  \emph{Sensors}, vol.~14, no.~2, pp. 3362--3394, 2014.

\bibitem{friedman2001greedy}
J.~H. Friedman, ``Greedy function approximation: a gradient boosting machine,''
  \emph{Annals of statistics}, pp. 1189--1232, 2001.

\bibitem{Chen:2016:XST:2939672.2939785}
T.~Chen \emph{et~al.}, ``{XGBoost}: A scalable tree boosting system,'' in
  \emph{Proceedings of the 22nd ACM SIGKDD International Conference on
  Knowledge Discovery and Data Mining}, ser. KDD '16.\hskip 1em plus 0.5em
  minus 0.4em\relax New York, NY, USA: ACM, 2016, pp. 785--794.

\bibitem{hochreiter1997long}
S.~Hochreiter \emph{et~al.}, ``Long short-term memory,'' \emph{Neural
  computation}, vol.~9, no.~8, pp. 1735--1780, 1997.

\bibitem{kidzinski_automatic_2019}
{\L}.~Kidzi{\'n}ski \emph{et~al.}, ``Automatic real-time gait event detection
  in children using deep neural networks,'' \emph{PloS one}, vol.~14, no.~1, p.
  e0211466, 2019.

\bibitem{Babrak2019}
L.~M. Babrak \emph{et~al.}, ``Traditional and digital biomarkers: Two worlds
  apart?'' \emph{Digital biomarkers}, vol.~3, no.~2, pp. 92--102, Aug. 2019.

\bibitem{Nagasaki2004AsymmetricVA}
H.~Nagasaki, ``Asymmetric velocity and acceleration profiles of human arm
  movements,'' \emph{Experimental Brain Research}, vol.~74, pp. 319--326, 2004.

\bibitem{sawyer_asymmetry_1993}
R.~N. Sawyer \emph{et~al.}, ``\BIBforeignlanguage{eng}{Asymmetry of forearm
  rolling as a sign of unilateral cerebral dysfunction},''
  \emph{\BIBforeignlanguage{eng}{Neurology}}, vol.~43, no.~8, pp. 1596--1598,
  Aug. 1993.

\bibitem{Pretegiani2017}
E.~Pretegiani \emph{et~al.}, ``Eye movements in {Parkinson}'s disease and
  inherited {Parkinsonian} syndromes,'' \emph{Frontiers in neurology}, vol.~8,
  pp. 592--592, Nov. 2017.

\bibitem{Jin2020}
B.~Jin \emph{et~al.}, ``Diagnosing {Parkinson} disease through facial
  expression recognition: Video analysis,'' \emph{Journal of medical Internet
  research}, vol.~22, no.~7, pp. e18\,697--e18\,697, Jul. 2020.

\bibitem{Rafayet2021}
M.~R. Ali \emph{et~al.}, ``Facial expressions can detect {P}arkinson's disease:
  preliminary evidence from videos collected online,'' \emph{npj Digital
  Medicine}, vol.~4, no.~1, p. 129, 2021.

\bibitem{Fabbri2017}
M.~Fabbri \emph{et~al.}, ``Speech and voice response to a levodopa challenge in
  late-stage {Parkinson}’s disease,'' \emph{Frontiers in Neurology}, vol.~8,
  p. 432, 2017.

\end{thebibliography}

\end{document}